\pdfoutput=1
\documentclass[conference]{IEEEtran}
\usepackage{times}
\usepackage[dvipsnames]{xcolor}

\usepackage[numbers]{natbib}
\usepackage{multicol}
\usepackage{wrapfig}
\usepackage{ulem}
\usepackage{amsmath, amsthm}

\usepackage{multicol}
\usepackage{amsfonts}
\usepackage{amssymb}

\usepackage{color}
\usepackage[ruled]{algorithm2e}   
\usepackage[pdftex]{graphicx}
\usepackage{booktabs}
\usepackage{silence}
\usepackage{utfsym}
\usepackage{dsfont}
\usepackage{mathabx}
\usepackage{tabularx}
\usepackage{graphicx}
\usepackage{float}
\usepackage{subfigure}
\usepackage{caption}
\usepackage{multirow}
\usepackage{hyperref}
\usepackage{colortbl}
\usepackage{enumerate}

\usepackage{amsmath}
\usepackage{array}
\usepackage{booktabs}
\usepackage{multicol,multirow}
\usepackage{setspace}
\usepackage{makecell}
\usepackage{pifont}
\usepackage{colortbl}
\usepackage{framed}
\usepackage{titletoc}
\usepackage{color}
\usepackage{caption, subcaption, overpic, textpos}
\usepackage{titlesec}
\usepackage{comment}
\usepackage{listings}

\definecolor{DeepBlue}{RGB}{65,100,170}
\newcommand{\underfigtab}{\vspace{-10pt}}

\begin{document}

\newcommand{\red}[1]{\textcolor{red}{#1}}
\newcommand{\blue}[1]{\textcolor{blue}{#1}}
\newcommand{\brown}[1]{\textcolor{brown}{#1}}
\newcommand{\wenye}[1]{\red{(Wenye: {#1})}}
\newcommand{\sizhe}[1]{\blue{(Sizhe: {#1})}}
\newcommand{\kerui}[1]{\brown{(Kerui: {#1})}}
\newcommand{\jiangmiao}[1]{\red{(Jiangmiao: {#1})}}
\title{Novel Demonstration Generation with Gaussian Splatting Enables Robust One-Shot Manipulation}





\author{Sizhe Yang$^{*,1,2}$\quad Wenye Yu$^{*,1,3}$\quad Jia Zeng$^{1}$\quad Jun Lv$^{3}$\quad Kerui Ren$^{1,3}$ \\ Cewu Lu$^{3}$\quad Dahua Lin$^{1,2}$\quad Jiangmiao Pang$^{1,\dagger}$ \\ 
$^{1}$Shanghai AI Laboratory 
\quad $^{2}$The Chinese University of Hong Kong \\
$^3$Shanghai Jiao Tong University \\
\textcolor{gray}{$^{*}$ Equal contributions \quad $^{\dagger}$ Corresponding author}\\
Project page: \textcolor{DeepBlue}{\url{https://yangsizhe.github.io/robosplat/}
}
}

\noindent
\twocolumn[{
\renewcommand\twocolumn[1][]{#1}
\maketitle
\vspace{-4mm}
\begin{center}
    \centering
    \captionsetup{type=figure}
    \includegraphics[width=1.00\textwidth]{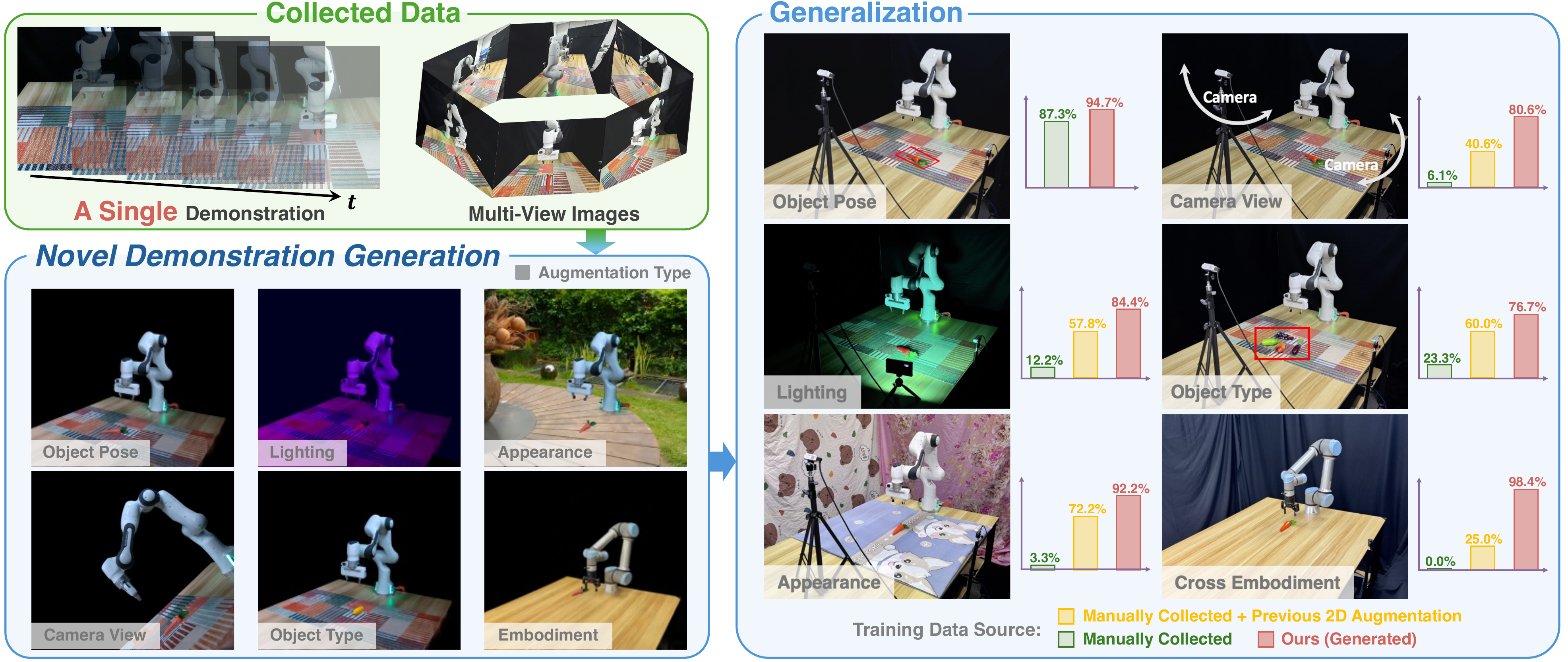}
    \captionof{figure}{{\textbf{Starting from a single expert demonstration and multi-view images, our method generates diverse and visually realistic data for policy learning, enabling robust performance across six types of generalization in the real world. Compared to previous 2D data augmentation methods, our approach achieves significantly better results across various generalization types. Notably, we achieve this within a unified framework.}} 
    \label{fig:teaser}
    \underfigtab
    }
    \vspace{2mm}
\end{center}
}]

\begin{abstract}
Visuomotor policies learned from teleoperated demonstrations face challenges such as lengthy data collection, high costs, and limited data diversity. Existing approaches address these issues by augmenting image observations in RGB space or employing Real-to-Sim-to-Real pipelines based on physical simulators. 
However, the former is constrained to 2D data augmentation, while the latter suffers from imprecise physical simulation caused by inaccurate geometric reconstruction.
This paper introduces RoboSplat, a novel method that generates diverse, visually realistic demonstrations by directly manipulating 3D Gaussians.
Specifically, we reconstruct the scene through 3D Gaussian Splatting (3DGS), directly edit the reconstructed scene, and augment data across six types of generalization with five techniques: 3D Gaussian replacement for varying object types, scene appearance, and robot embodiments; equivariant transformations for different object poses; visual attribute editing for various lighting conditions; novel view synthesis for new camera perspectives; and 3D content generation for diverse object types.
Comprehensive real-world experiments demonstrate that RoboSplat significantly enhances the generalization of visuomotor policies under diverse disturbances. Notably, while policies trained on hundreds of real-world demonstrations with additional 2D data augmentation achieve an average success rate of 57.2\%, RoboSplat attains 87.8\% in one-shot settings across six types of generalization in the real world.

\end{abstract}

\IEEEpeerreviewmaketitle

\section{Introduction}
\label{sec:intro}

Imitation learning for visuomotor policies has emerged as a promising paradigm in robot manipulation. However, policies learned through imitation often display limited robustness in deployment scenarios that differ substantially from expert demonstrations, primarily due to insufficient coverage of visual domains in the training data. Increasing the volume and diversity of real-world data is an effective strategy for enhancing robustness~\cite{chi2024universal}; however, acquiring human-collected demonstrations is prohibitively time-consuming and labor-intensive.
Consequently, substantial efforts have been devoted to generating diverse expert data without engaging with real-world environments~\cite{yuan2025roboengine, yuan2024learningmanipulateanywherevisual, singh2024dextrah, chen2024roviaugrobotviewpointaugmentation, chen2024semanticallycontrollableaugmentationsgeneralizable, yu2023scalingrobotlearningsemantically, chen2023genaug, mandi2023cactiframeworkscalablemultitask, tian2024view, xue2025demogen}. 

Simulated environments offer a low-cost platform for data synthesis~\cite{singh2024dextrah, yuan2024learningmanipulateanywherevisual}.
However, the Sim-to-Real gap presents significant challenges that hinder policy performance in real-world scenarios. 
Although Real-to-Sim-to-Real pipelines can narrow this gap considerably, replicating real-world manipulation scenes in simulation remains complex and labor-intensive. In particular, inaccuracies in geometric reconstructions often lead to imprecise physical simulations. 
Moreover, existing Real-to-Sim-to-Real approaches primarily generate data within monotonously reconstructed scenes, resulting in policies that are tailored only to those specific environments. 
Another line of work sheds light on augmenting image observations for better visual generalization.
By editing different semantic parts of the image, these approaches generate novel scene configurations, in terms of background appearances~\cite{yuan2025roboengine, chen2023genaug, yu2023scalingrobotlearningsemantically, chen2024semanticallycontrollableaugmentationsgeneralizable}, embodiment types~\cite{chen2024roviaugrobotviewpointaugmentation}, object types~\cite{yu2023scalingrobotlearningsemantically}, and camera views~\cite{tian2024view}.
While these image augmentation methods are convenient, their limited consideration of 3D spatial information results in spatially inaccurate data generation.
For more effective data augmentation, explicit 3D representations that retain accurate spatial information and are realistically renderable are required.

Recently, 3D Gaussian Splatting (3DGS)~\cite{kerbl20233dgaussiansplattingrealtime} has become a burgeoning approach to superior reconstruction and rendering.
Thanks to its explicit representation of the scene, 3DGS enables interpretable editing of the reconstructed scene, which paves the way for generating novel manipulation configurations. 
Furthermore, as a 3D representation of the scene, 3DGS retains spatial information from the real world and allows for consistent rendering from multiple perspectives, which makes it the real-world counterpart of a simulator's graphics engine for generating novel demonstrations.

Based on that, we propose \textbf{RoboSplat}, a novel and efficacious approach to demonstration generation with Gaussian Splatting. 
Empowered by 3DGS, we achieve a high-fidelity reconstruction of the manipulation scene.
In order to align the reconstructed scene with real-world counterparts, we devise a novel frame alignment pipeline leveraging differentiable rendering of Gaussian Splatting.
3D Gaussians of different scene components are segmented using off-the-shelf segmentation models and the robot United Robotics Description Format (URDF).
Remarkably, as illustrated in Fig.~\ref{fig:teaser}, a single collected expert trajectory enables us to generate novel demonstrations across a wide range of visual domains.
To be specific, RoboSplat augments data across six types of generalization with five techniques: 3D Gaussian replacement for varying object types, scene appearance, and robot embodiments; equivariant transformations for different object poses; visual attribute editing for various lighting conditions; novel view synthesis for new camera perspectives; and 3D content generation for diverse object types.

Compared to previous Real-to-Sim-to-Real and image augmentation approaches, RoboSplat achieves more diverse and spatially accurate data generation.
Extensive real-world experiments demonstrate that RoboSplat significantly enhances the robustness of visuomotor policies against multiple disturbances across tasks involving pick and place, tool use, functional motion, articulated object manipulation, and long-horizon skills.
Specifically, compared to policies trained on hundreds of real-world demonstrations that are further enriched with 2D data augmentation, our method increases the average success rate from 57.2\% to 87.8\%.



\begin{figure*}[ht]
    \centering
    \includegraphics[width=0.98\linewidth]{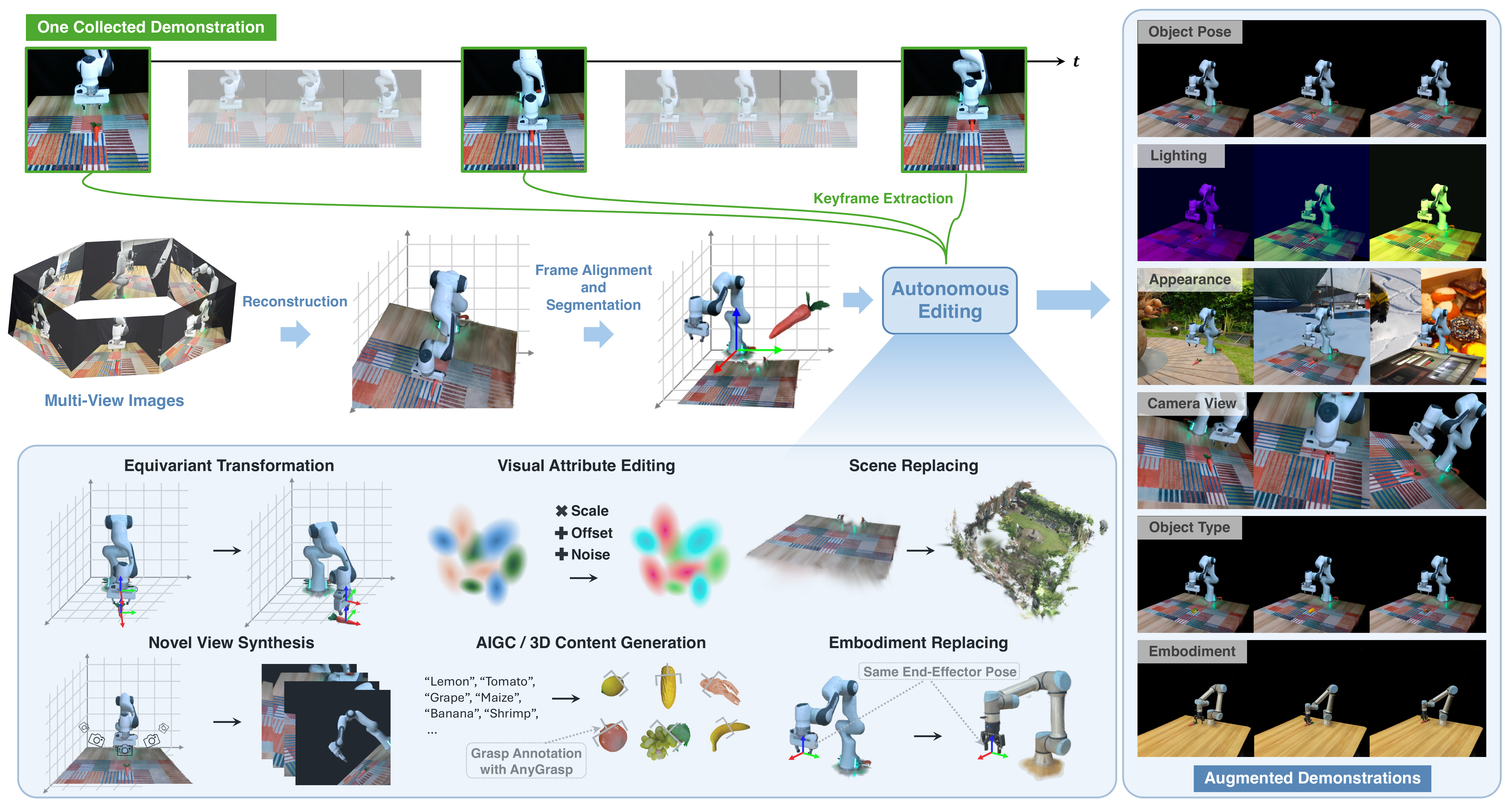}
    \caption{\textbf{Method overview.} We start from a single manually collected demonstration and multi-view images that capture the whole scene. The former provides task-related keyframes, while the latter helps scene reconstruction. After aligning the reconstructed frame with the real-world frame and segmenting different scene components, we carry out autonomous editing of the scene in pursuit of six types of augmentation.}
    \label{fig:method}
    \vspace{-3mm}
\end{figure*}

\section{Related work}
\label{sec:related_work}

\subsection{Generalizable Policy in Robot Manipulation}

Recent advancements in manipulation have significantly enhanced generalization.
Some studies design the policy architecture to endow it with equivariant properties, which is helpful to generalizing to different object poses~\cite{yang2024equibotsim3equivariantdiffusionpolicy, yang2024equivactsim3equivariantvisuomotorpolicies, ryu2023equivariantdescriptorfieldsse3equivariant, chun2023localneuraldescriptorfields}.
One-shot imitation learning approaches like~\cite{vosylius2024instantpolicyincontextimitation,simeonov2021neuraldescriptorfieldsse3equivariant, biza2023oneshotimitationlearninginteraction, vitiello2023oneshotimitationlearningpose, zhang2024oneshotimitationlearninginvariance} enable the policy to handle various object poses given only one demonstration. 
Furthermore, some other work focuses on generalizing the policy to different camera views~\cite{yuan2024learningmanipulateanywherevisual, mvmwm, movie}, scene appearance~\cite{levy2024p3, seer}, and embodiments~\cite{chi2024universal}.
Some studies exploit the power of Large Language Models (LLMs) and Vision Language Models (VLMs) to endow robots with generalization abilities~\cite{irpan2022can, brohan2023rt2visionlanguageactionmodelstransfer, embodimentcollaboration2024openxembodimentroboticlearning, dalal2024local}. 
Instead of adopting generalizable policy architecture, auxiliary learning objectives and powerful foundation models, our work is concentrated on generating high-quality, diverse, and realistic data to instill generalization abilities to the learned policy.

\subsection{Data Augmentation for Policy Learning}

Given limited training data, data augmentation emerges as a way to improve the robustness of the policy. 
Previous work adopts image augmentation techniques to improve the resistance of visuomotor policies to observation noises~\cite{laskin2020reinforcementlearningaugmenteddata, kostrikov2021imageaugmentationneedregularizing, mandlekar2021matterslearningofflinehuman, mandlekar2023mimicgendatagenerationscalable, fan2021secantselfexpertcloningzeroshot, hansen2021generalizationreinforcementlearningsoft, hansen2021stabilizingdeepqlearningconvnets}.
However, these methods are mainly evaluated in simulated environments. 
To deploy learned policies in real-world setting, some previous work focuses on augmenting the appearance of the scene by incorporating image-inpainting models~\cite{yu2023scalingrobotlearningsemantically, chen2024semanticallycontrollableaugmentationsgeneralizable, chen2023genaug, mandi2023cactiframeworkscalablemultitask}. 
Moreover,~\citet{tian2024view} generate augmented task demonstrations from different camera views and aim to learn a view-invariant policy.
\citet{ameperosa2024rocoda}.
\citet{chen2024roviaugrobotviewpointaugmentation} further devise a cross-embodiment pipeline by inpainting different robots to image observations. 
Nonetheless, these studies mainly augment task demonstrations on 2D images, which lack spatial information. Hence, only limited augmentation can be achieved, and the augmented demonstrations might be unrealistic compared to those generated directly from 3D representations.
Our work reconstructs the scene with 3D Gaussian Splatting and edits the 3D representation for data augmentation, enabling our policy to achieve comprehensive generalization across object poses, object types, camera views, lighting conditions, scene appearance, and various embodiments..

\subsection{Gaussian Splatting in Robotics}

3D Gaussian Splatting (3DGS)~\cite{kerbl20233dgaussiansplattingrealtime} serves as an explicit radiance field representation for real-time rendering of 3D scenes.
Previous work leverages 3DGS to select proper grasp poses~\cite{ji2024graspsplatsefficientmanipulation3d, zheng2024gaussiangrasper3dlanguagegaussian}. 
Furthermore,~\citet{lu2024manigaussiandynamicgaussiansplatting} exploit 3DGS to construct dynamics of the scene for multi-task robot manipulation. 
In order to predict the consequence of robots' interactions with the environment,~\citet{shorinwa2024splatmovermultistageopenvocabularyrobotic} leverage 3D semantic masking and infilling to visualize the motions of the objects that result from the interactions. 
Another line of work adopts the Real-to-Sim-to-Real pipeline, and utilizes 3DGS to reconstruct the real-world scene~\cite{li2024robogsimreal2sim2realroboticgaussian, qureshi2024splatsimzeroshotsim2realtransfer, wu2024rlgsbridge3dgaussiansplatting, torne2024reconcilingrealitysimulationrealtosimtoreal}. However, importing reconstructed real-world objects to simulation is a strenuous process, and physical interactions tend to suffer from large sim-to-real gaps due to the flawed geometric reconstruction and lack of physical information in 3D reconstruction.
Some recent work on 3DGS is centered around editing and relighting of the scene~\cite{ye2024gaussiangroupingsegmentedit, liang2024gsir3dgaussiansplatting,gao2024relightable3dgaussiansrealistic}. 
Our method enables autonomous editing of the reconstructed scene to generate diverse demonstrations with various configurations.

\section{Preliminaries}
\label{sec:preliminaries}

3D Gaussian Splatting (3DGS)~\cite{kerbl20233dgaussiansplattingrealtime} utilizes multi-view images for high-fidelity scene reconstruction. The scene is represented by a set of Gaussians $\{g_i\}_{i=1}^{N}$, where each Gaussian $g_i$ consists of a position vector $\mu_i\in \mathbb{R}^3$, a rotation matrix $R_i\in \mathbb{R}^{3\times3}$, a scaling matrix $S_i=diag(s)(s\in \mathbb{R}^3)$,
an opacity factor $\alpha_i\in\mathbb{R}$, 
and spherical harmonic coefficients $c_i$ that encapsulate the view-dependent color appearance of the Gaussian.
Given the scaling matrix and rotation matrix, the covariance matrix $\Sigma_i$ is calculated as follows:

\begin{equation*}
    \Sigma_i = R_iS_iS_i^\top R_i^\top.
\end{equation*}

To derive the color $C$ of a particular pixel during rendering procedure, 3DGS exploits a typical neural point-based approach, similar to~\citet{kopanas2022neural}, where the final color value is calculated as follows:

\begin{align*}
    C &= \sum\limits_{i=1}^{N} c_i o_i \prod\limits_{j=1}^{j=i-1} (1-o_j), \\
    o_i &= \alpha_i \cdot \exp{(\frac{1}{2}\delta_i^\intercal \Sigma_{i,2D}^{-1}\delta_i)},
\end{align*}

where $N$ is the number of Gaussians that overlap with the pixel. Besides, $\alpha_i$ denotes the opacity of the $i$-th Gaussian. $\delta_i\in \mathbb{R}^{2}$ denotes the offset between the current pixel and the center of the $i$-th Gaussian projected to 2D image. $\Sigma_{i,2D}\in \mathbb{R}^{2\times2}$ stands for the covariance matrix of the $i$-th Gaussian projected to 2D image.

\section{Methodology}
\label{sec:method}

To generate high-fidelity and diverse data from a single expert trajectory, we present RoboSplat, a novel demonstration generation approach based on 3DGS. 
An overview of our method is shown in Fig.~\ref{fig:method}.
In this section, we describe RoboSplat in detail.
We begin with the process of reconstruction and preprocessing in Sec.~\ref{sec:scene_reconstruction}, 
which includes object and scene reconstruction, frame alignment with differentiable rendering, and novel pose generation for the robot and objects.
With all the Gaussian models ready, we generate novel demonstrations and perform data augmentation in terms of object poses, object types, camera views, scene appearance, lighting conditions, and embodiments, as described in Sec.~\ref{sec:data_augmentation}. 
Finally, a visuomotor policy is trained on the augmented demonstrations and directly deployed on real robots, as detailed in Sec.~\ref{sec:policy}.




\subsection{Reconstruction and Preprocessing}
\label{sec:scene_reconstruction}

In pursuit of a high-fidelity reconstruction of the scene, we first capture a set of RGB images whose corresponding viewpoints should be as various as possible. During this process, the scene remains static and the robot is fixed at its default joint configuration, which we refer to as $q_{\text{default}}$. With the images ready, we utilize COLMAP~\cite{schoenberger2016mvs, schoenberger2016sfm} to obtain a sparse scene reconstruction and an estimation of the camera pose corresponding to each image. To further enhance the reconstruction precision, we gain an depth estimation for each image with Depth Anything~\cite{yang2024depth}. The images, camera poses, and depth prior  serve as inputs to 3DGS~\cite{kerbl20233dgaussiansplattingrealtime}, which returns 3D Gaussians representing the entire scene $\mathcal{G}_{\text{scene}}$, which contains 3D Gaussians corresponding to the robot, dubbed $\mathcal{G}_{\text{robot}}$. 

However, the reconstructed 3D Gaussians of the robot are represented in an arbitrary frame $\mathcal{F}_{\text{scene}}$, and hence we need to align it with the real-world coordinate frame $\mathcal{F}_{\text{real}}$ to facilitate automated editing.

\begin{figure}[t!]
    \centering
    \includegraphics[width=0.8\linewidth]{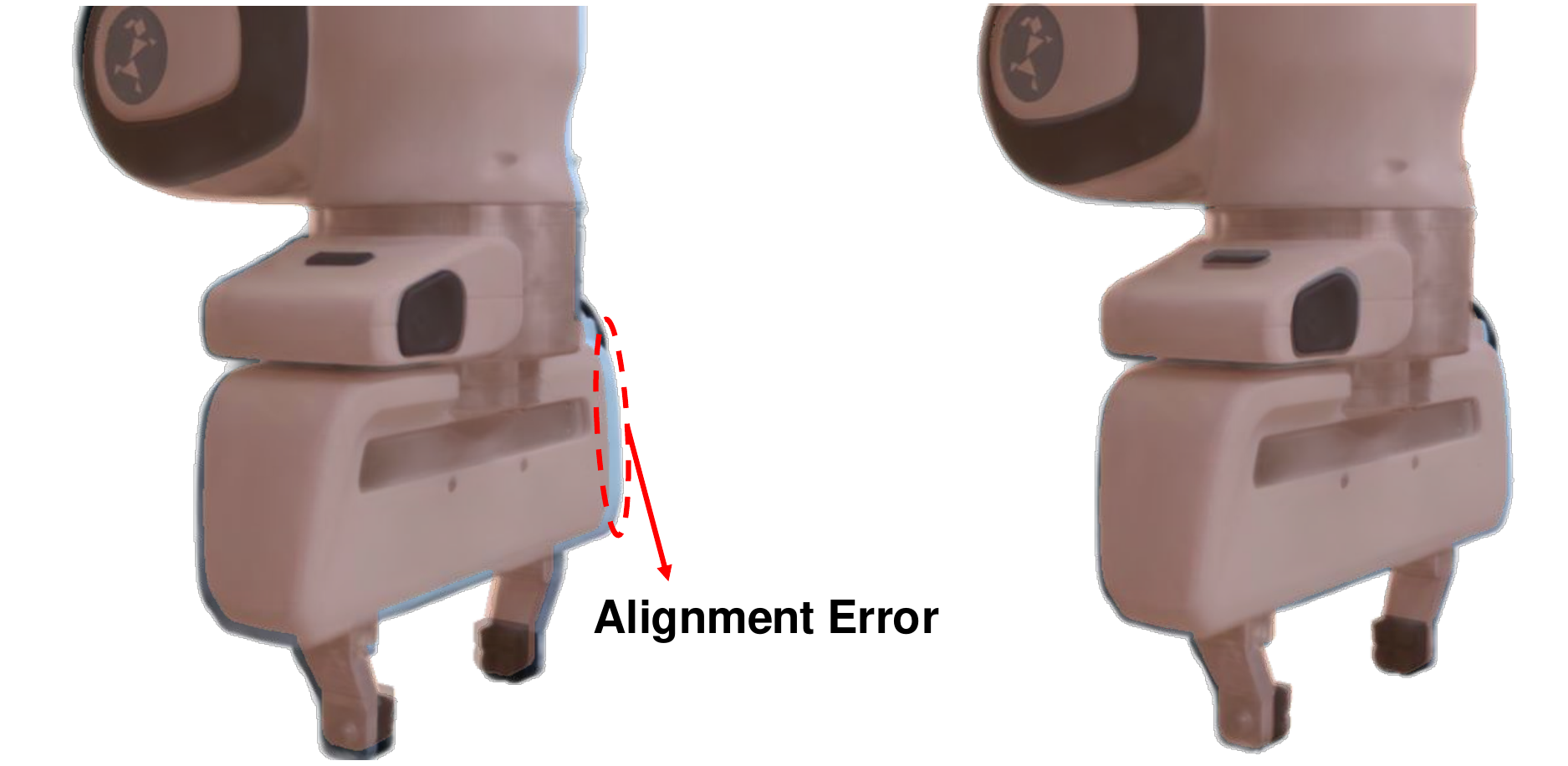}
    \caption{\textbf{Comparison of frame alignment results between ICP and fine-grained optimization with differentiable rendering.} The semi-transparent orange overlay represents the ground truth rendered with URDF from the same camera view. The \textbf{left} shows the results of ICP, which have larger errors, while the \textbf{right} shows the results after further fine-grained optimization using differentiable rendering.}
    \label{fig:diff_align_compare}
\end{figure}

\begin{figure}[t!]
    \centering
    \includegraphics[width=\linewidth]{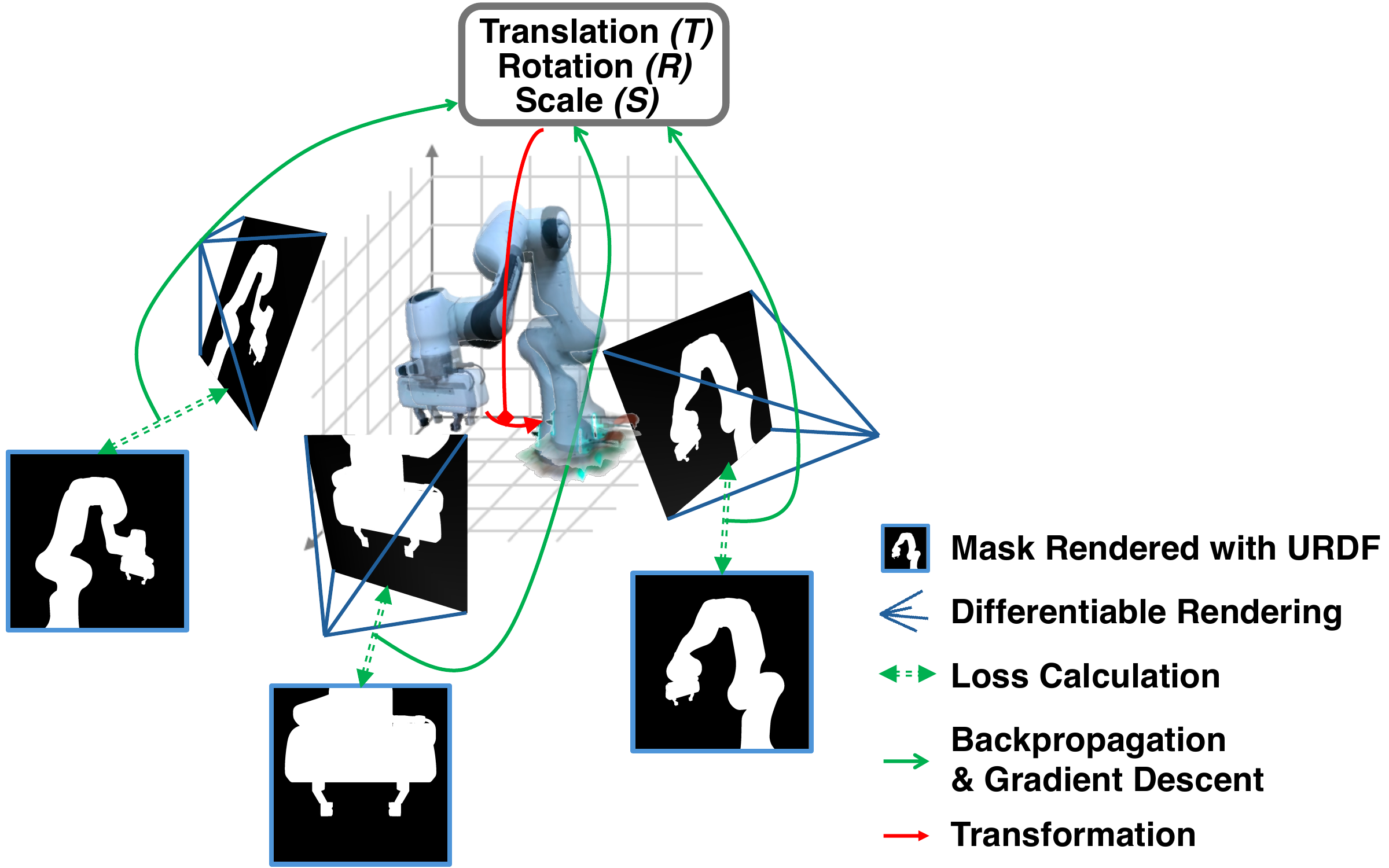}
    \caption{\textbf{Illustration of frame alignment with differentiable rendering.} The loss is calculated between the mask rendered using Gaussian Splatting and the mask rendered with URDF. Subsequently, backpropagation and gradient descent are used to optimize the translation, rotation, and scale, which are then applied to the 3D Gaussians.}
    \label{fig:diff_align}
    \vspace{-6mm}
\end{figure}

The robot URDF gives us access to the robot base frame $\mathcal{F}_{\text{URDF}}$. The real-world robot frame $\mathcal{F}_{\text{robot}}$, $\mathcal{F}_{\text{URDF}}$, and $\mathcal{F}_{\text{real}}$ are all aligned with each other.
Hence, the actual problem turns into the frame alignment from $\mathcal{F}_{\text{scene}}$ to $\mathcal{F}_{\text{URDF}}$. We denote the transformation matrix as $\mathcal{T}_{\text{URDF, scene}}$. 
While point cloud registration approachs, such as Iterative Closest Point (ICP)~\cite{besl1992method},  serve as a common solution to it, we find that there is still major misalignment between the two frames aligned with point cloud registration, as illustrated in Fig.~\ref{fig:diff_align_compare}. The reason lies in the fact that point cloud registration is based on point coordinates, whereas 3D Gaussians have a scale attribute, which causes a mismatch between point coordinates and the appearance. Therefore, we exploit the differentiable rendering of 3DGS to do further fine-grained alignment, as depicted in Fig.~\ref{fig:diff_align}.

Suppose $\hat{\mathcal{T}}_{\text{URDF, scene}}^{0}$ is the initial transformation matrix obtained through ICP. We first apply $\hat{\mathcal{T}}_{\text{URDF, scene}}^{0}$ to $\mathcal{G}_{\text{robot}}$, leading to a partially aligned robot Gaussian $\hat{\mathcal{G}}_{\text{robot}}$.
The aim of further alignment is to derive another transformation matrix $\hat{\mathcal{T}}_{\text{rel}}$, such that applying $\hat{\mathcal{T}}_{\text{rel}}$ to $\hat{\mathcal{G}}_{\text{robot}}$ gives a better alignment to the pose of the robot defined in URDF. 
For this sake, we select $N$ canonical camera views to capture the segmentation masks $\{\mathcal{I}_i^{\text{URDF}}\}_{i=1}^{N}$ and $\{\mathcal{I}_i^{\text{Gaussian}}\}_{i=1}^{N}$ (the pixel value is 1 if it belongs to the robot; otherwise, it is 0) with the robot URDF and $\hat{\mathcal{G}}_{\text{robot}}$ respectively. The pixel-wise differences between the images from the same canonical views are averaged to form the objective function of alignment:

\begin{equation*}
    \mathcal{L}_{\text{align}} = \frac{1}{N} \sum_{i=1}^N (\mathcal{I}_i^{\text{URDF}} - \mathcal{I}_i^{\text{Gaussian}})^2.
\end{equation*}

Due to the differentiability of Gaussian Splatting, we can rewrite the objective function as $\mathcal{L}_{\text{align}}(\hat{\mathcal{T}}_{\text{rel}})$ and optimize $\hat{\mathcal{T}}_{\text{rel}}$ through gradient descent. 
The optimized $\hat{\mathcal{T}}_{\text{rel}}$ is composed with $\hat{\mathcal{T}}_{\text{URDF, scene}}^{0}$, the result of which is applied to $\mathcal{G}_{\text{scene}}$ to form the scene reconstruction in $\mathcal{F}_{\text{real}}$. 
We refer to the aligned 3D Gaussians as $\mathcal{G}_{\text{scene}}^{\ast}$.

In order to decompose the scene into different parts, we first leverage Grounded-SAM~\cite{ren2024grounded} to perform task-related object segmentation. 
Then, the masked images are used to reconstruct 3D Gaussians for the objects. 
The 3D Gaussians corresponding to each link of the robot are segmented using the point cloud of each link in $\mathcal{F}_{\text{URDF}}$, which can be obtained with the robot's URDF and the renderer. 
Specifically, if the position of a 3D Gaussian is within a threshold distance from the point cloud of a link, the 3D Gaussian is assigned to that link. 
If a 3D Gaussian does not belong to any object or any link of the robot, it is classified as background.
We suppose that the robot has $l$ links and there are totally $k$ objects in the scene. 
The reconstructed robot links, objects, and background are denoted as $\mathcal{G}_{\text{robot}}^{\ast}=\{\mathcal{G}_{\text{robot},i}^{\ast}\}_{i=1}^{l}$, $\mathcal{G}_{\text{obj}}^{\ast}=\{\mathcal{G}_{\text{obj},j}^{\ast}\}_{j=1}^{k}$, and $\mathcal{G}_{\text{bg}}^{\ast}$ respectively.

Similar to our frame alignment strategy, we utilize differentiable rendering to estimate the deployed camera poses in order to narrow the gap between the generated data and the deployment environment. The camera extrinsics are optimized through gradient descent, with the optimization objective:

\begin{equation*}
    \mathcal{L}_{\text{camera}} = SSIM(\mathcal{I}_{\text{Expert}}, \mathcal{I}_{\text{Gaussian}})^2,
\end{equation*}

where $\mathcal{I}_{\text{Expert}}$ denotes the image obtained from the collected expert demonstration, $\mathcal{I}_{\text{Gaussian}}$ represents the rendered image with reconstructed 3D Gaussians, and SSIM refers to Structural Similarity, which measures the perceptual similarity between two images.


Nonetheless, before moving on to novel demonstration generation, we need to figure out how to generate 3D Gaussians for the robot under novel joint configurations. To achieve that, we leverage the link-wise Gaussians $\{\mathcal{G}_{\text{robot},i}^{\ast}\}_{i=1}^{l}$ and the default joint configuration $q_{\text{default}}$. For each link $1\leq i\leq l$, we access its relative pose to robot base frame under arbitrary joint configuration $q$ through forward kinematics, denoted as $\mathcal{T}_{\text{fk}}^{i}(q)$. Hence, by transforming each link $i$ with $\mathcal{T}_{\text{fk}}^{i}(q)\mathcal{T}_{\text{fk}}^{i}(q_{\text{default}})^{-1}$, we derive the corresponding 3D Gaussians under configuration $q$. The entire 3D Gaussians are thereby derived by composing Gaussians of all $l$ links.
As for the manipulated objects, we apply transformations in a similar manner. The way 3D Gaussians are transformed is detailed in Appendix~\ref{subsec:applying_transformation}.

\subsection{Novel Demonstration Generation}
\label{sec:data_augmentation}


Utilizing 3D Gaussians in $\mathcal{F}_{\text{real}}$, we implement our demonstration augmentation process, which systematically enhances the expert demonstration $\mathcal{D}_{\text{expert}}$ across six aspects: 
object poses, object types, camera views, embodiment types, scene appearance, and lighting conditions.

\subsubsection{Object Pose}
\label{subsec:object_pose}

To perform object pose augmentation, we first extract keyframes from the expert demonstration using a heuristic approach. 
Whenever the gripper action toggles or joint velocities approach zero, we consider the current time step as a keyframe and record the end-effector pose with respect to robot base frame. 
After that, we apply rigid transformations to the target objects that are involved in the expert demonstration. 
The end-effector poses at keyframes are transformed equivariantly according to the target object. 
Eventually, we generate trajectories between consecutive keyframe poses with motion planning, the combination of which makes a complete augmented demonstration with novel object poses.

\subsubsection{Object Type}
\label{subsec:object_type}

The object types can be augmented with 3D Content Generation. 
We first prompt GPT-4~\cite{gpt4} to generate approximately 50 names of objects that can be grasped. 
Then, we use these object names as prompts to generate corresponding 3D Gaussians with a 3D content generation model~\cite{xiang2024structured}.
We utilize an off-the-shelf grasping algorithm~\cite{fang2023anygrasp} to generate grasp poses with respect to the object frame.
As we generate different object poses for augmentation, we obtain the corresponding end-effector poses by composing object pose and the grasp pose relative to the object, which turn into the keyframe poses in new demonstrations. 
The entire augmented trajectory is generated in the same manner as~\ref{subsec:object_pose}. 

\subsubsection{Camera View}
\label{subsec:camera_view}

One merit of 3DGS lies in its ability to perform novel view synthesis. Thereby, we are able to choose different camera poses from $\mathcal{D}_{\text{expert}}$ and obtain novel-view demonstrations. 
Although we can render novel-view observations from arbitrary camera pose, we need to ensure that the augmented camera view does not deviate so much from the expert that it loses sight of the manipulation scene. 
Hence, we first designate a target point $O_c=(x_c, y_c, z_c)$ in $\mathcal{F}_{\text{real}}$, towards which the camera should face during the entire episode. We then define a coordinate frame $\mathcal{F}_c$, whose origin is $O_c$ and orientation is the same as $\mathcal{F}_{\text{real}}$. 
The position of camera is represented by spherical coordinates $(r, \theta, \varphi)$ in $\mathcal{F}_c$. Thus, by limiting the target point within the manipulation scene and randomizing the spherical coordinates, we are able to generate camera poses that produce meaningful observations yet possess diversity.
The hyperparameters of randomization for the target point and the spherical coordinates are detailed in Appendix~\ref{subsec:aug_detail}.

\subsubsection{Embodiment Type}
\label{subsec:embodiment}

To generalize the expert demonstration to different types of robots, we replace $\mathcal{G}_{\text{robot}}^{\ast}$ with the 3D Gaussians of another embodiment, dubbed $\mathcal{G}_{\text{robot}}^{\text{new}}$, which is attained from the corresponding URDF file or real-world reconstruction. 
The keyframe end-effector poses are reused because they are embodiment-agnostic action representations.
Hence, through motion planning, we can easily derive the end-effector poses and joint positions of the new embodiment for all time steps in augmented demonstrations. 
The 3D Gaussians of the new embodiment under novel joint configurations is obtained from $\mathcal{G}_{\text{robot}}^{\text{new}}$ as mentioned in Sec.~\ref{sec:scene_reconstruction}. The policy trained on these augmented demonstrations is directly deployed on novel embodiments.

\subsubsection{Scene Appearance}
\label{subsec:appearance}

Inconsistency between scene appearance accounts for a large visual gap between training and deployment environments. 
To resolve this issue, we propose to exploit reconstructed diverse 3D scenes and also large-scale image datasets to augment the scene appearance.
We adopt COCO~\cite{lin2014microsoft} as the image dataset, and attach images to the table top and background 3D Gaussian planes that surround the entire manipulation scene. 
Moreover, we gather datasets for 3D reconstruction~\cite{hedman2018deep,yeshwanth2023scannet++,Knapitsch2017,barron2022mip}, and derive corresponding 3D Gaussians by 3DGS training. 
The resulting 3D Gaussian scenes substitute for $\mathcal{G}_{\text{bg}}^{\ast}$, forming novel scene appearance for data augmentation. 
The edge of utilizing reconstructed 3D scenes is their consistent and diverse geometry across multiple camera views, which helps produce more realistic demonstrations.
Nevertheless, due to the expense of 3DGS training on large-scale reconstruction datasets, we complement them with 2D images for greater appearance diversity.

\subsubsection{Lighting Condition}
\label{subsec:lighting}

Discrepancy in lighting conditions is another barrier to deploying trained policy in unseen scenarios. To compensate for that, we augment the diffuse color of each Gaussian in the reconstructed scene through random scaling, offset, and noise. 
Concretely, for a Gaussian with original diffuse color $(r, g, b)$, the augmented diffuse color values can be expressed as $(s_r r + o_r + \Delta_r, s_g g+o_g+\Delta_g, s_b b + o_b + \Delta_b)$, where $(s_r, s_g, s_b)$ stand for scaling factors, $(o_r, o_g, o_b)$ stand for offsets, and $(\Delta_r, \Delta_g, \Delta_b)$ stand for random Gaussian noise. 
The scaling factors and offsets simulate changes in color contrast and scene brightness. Thus, they are shared among all the Gaussians in the scene.
On the other hand, the random Gaussian noise is sampled independently for each Gaussian to simulate noise in images captured by cameras.
The details of scaling factors, offsets, and Gaussian noise are elaborated in Appendix~\ref{subsec:aug_detail}.

An illustration of augmented demonstrations with six types of generalizations can be found in Appendix~\ref{subsec:aug_detail}.

\subsection{Policy Training}
\label{sec:policy}

We employ a modern, widely adopted transformer-based architecture~\cite{haldar2024baku, seer, octo, gr1} to serve as the policy network, which is detailed in Appendix~\ref{subsec:policy_architecture}.
We process RGB images with ResNet-18~\cite{resnet}, and encode joint state using a multi-layer perceptron (MLP). The latent of images and robot state is fed into a transformer encoder. Finally, an action decoder utilizes an MLP to convert the action latent into the action vector $a_t$.
The policy is trained with Behavioural Cloning (BC) in an end-to-end manner, aiming to maximize the likelihood of expert actions in demonstrations. 
We denote $o_k \triangleq (I_k, q_k)$ as the observation at the $k$-th frame of demonstrations $\mathcal{D}$, and $\pi$ as our policy.
The loss function can then be expressed as

\begin{equation*}
    \mathcal{L}^{\text{BC}} = \mathbb{E}_{(o_k, a_k)\sim \mathcal{D}}\|a_k - \pi(o_k)\|^2.
    \label{eq:BC}
\end{equation*}

Specifically, $I_k$ consists of two images from different eye-on-base cameras.
We adopt relative end-effector pose as the action representation, which depicts the relative transformation between two consecutive end-effector poses under robot base frame.
Further details of the training process can be found in Appendix~\ref{subsec:training_details}.

\begin{figure}[h!]
    \vspace{0mm}
    \centering
    \includegraphics[width=0.95\linewidth]{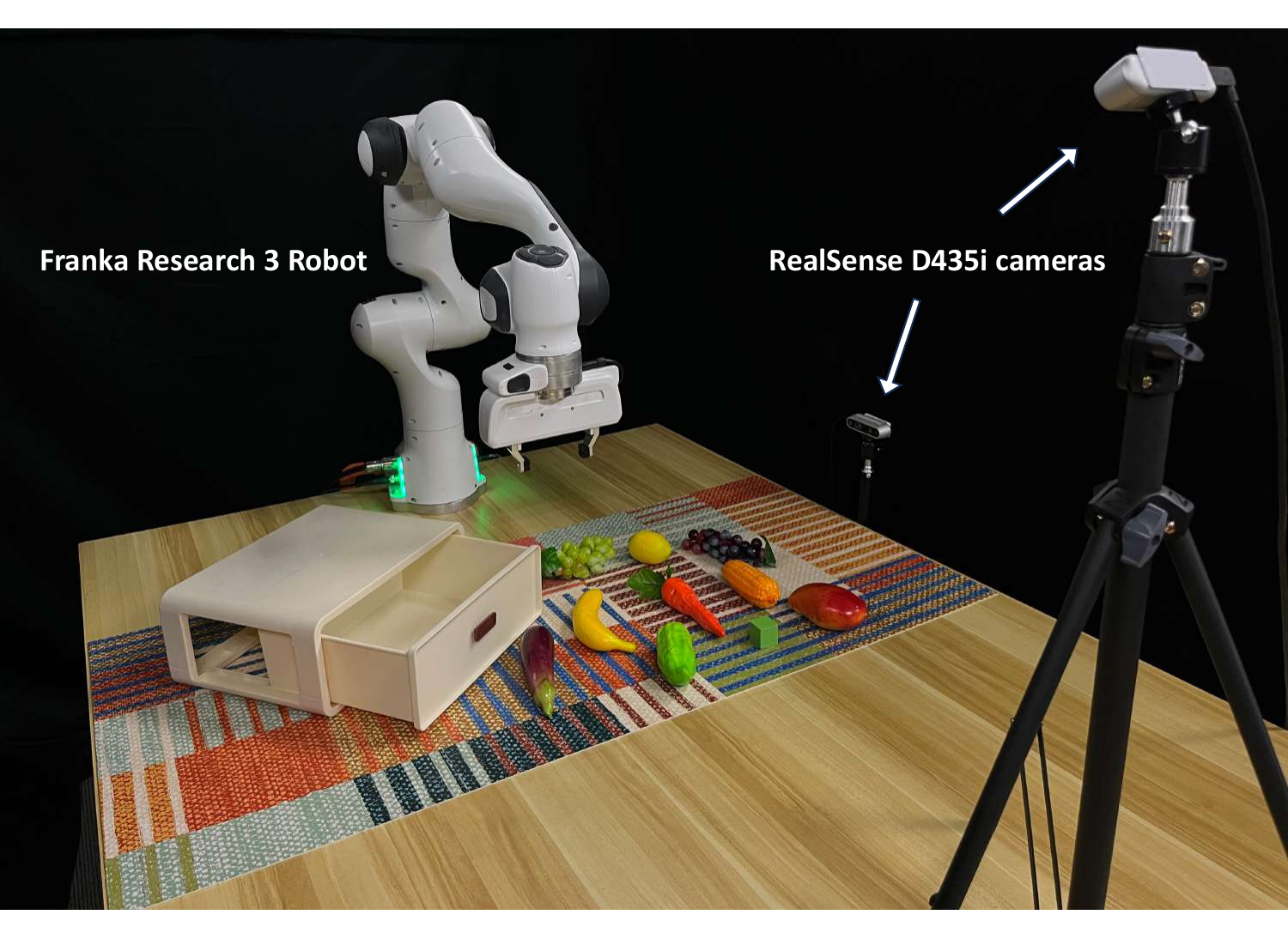}
    \vspace{-2mm}
    \caption{\textbf{Real-world experiment setup.} We employ a Franka Research 3 Robot and two eye-on-base RealSense D435i cameras.}
    \label{fig:real_setup}
    \vspace{-4mm}
\end{figure}

\section{Experiments}
\label{sec:experiments}

\begin{figure*}[h!]
    \centering
    \includegraphics[width=\linewidth]{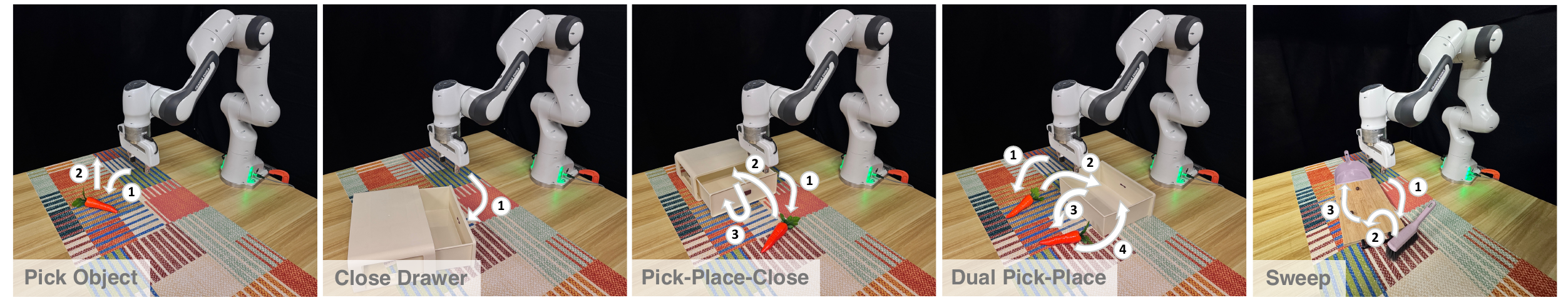}
    \caption{\textbf{Task illustration.} We design five manipulation tasks for real-world evaluation: \textit{Pick Object}, \textit{Close Drawer}, \textit{Pick-Place-Close}, \textit{Dual Pick-Place} and \textit{Sweep}, whose details are elaborated in Sec.~\ref{sec:exp_setup}.}
    \label{fig:task_setup}
    \vspace{-5mm}
\end{figure*}


We conduct comprehensive experiments in the real world to verify the effectiveness of our demonstration generation pipeline. Specifically, we aim to answer: given a single expert demonstration and multi-view images of the scene, 

1) How efficient is data generation compared to manually collecting data?

2) How does the policy trained on generated demonstrations perform across various tasks compared to that trained on manually collected data? 

3) How does the policy perform as the generated data scale up?

4) Can generated demonstrations enhance the robustness of the policy when facing various deployment settings, such as changes in object types, camera views, scene appearance, lighting conditions, and embodiment types? 

\subsection{Experimental Setup}
\label{sec:exp_setup}

The real-world experiment setup is presented in Fig.~\ref{fig:real_setup}.
Concretely, we collect the expert demonstration on Franka Research 3 (FR3) Robot. Two Intel Realsense D435i eye-on-base cameras are mounted on the table top, capturing RGB image observations for the policy. 
We employ a 3D SpaceMouse to collect teleoperated demonstrations at a frequency of 10 Hz. 
Policy inference is carried out on an NVIDIA RTX4090 GPU, with a latency of 0.1s imposed.

In order to manifest the generalization ability of our pipeline to different task settings, we select five tasks for evaluation: \textit{Pick Object}, \textit{Close Drawer}, \textit{Pick-Place-Close}, \textit{Dual Pick-Place}, and \textit{Sweep}. 

In \textit{Pick Object} task, the policy picks up a target object which is placed at different poses within a 30cm$\times$40cm workspace.
In \textit{Close Drawer} task, the policy closes a drawer whose position is constrained to a 15cm$\times$40cm workspace, while its rotation about the z-axis is restricted to $[-\frac{\pi}{8}, \frac{\pi}{8}]$.
In \textit{Pick-Place-Close} task, the policy is expected to grasp an object, place it in the drawer, and then close the drawer. 
The drawer is placed in a 5cm$\times$5cm workspace, with a fixed orientation.
The target object is located in a 10cm$\times$10cm workspace, whose rotation falls into range $[-\frac{\pi}{8}, \frac{\pi}{8}]$.
In \textit{Dual Pick-Place} task, the policy attempts to pick two target objects in a row and place them in a fixed drawer.
Both of the objects are located in 10cm$\times$10cm workspaces, with yaw angles between $-\frac{\pi}{8}$ and $\frac{\pi}{8}$.
In \textit{Sweep} task, the robot should first pick up a broom and then sweeps the chocolate beans into a dustpan. The broom is randomly placed within a 10cm × 10cm area, and the chocolate beans are randomly placed on the chopping board.
Task setups are illustrated in Fig.~\ref{fig:task_setup}.
These five tasks require proficiency in executing basic pick-and-place actions, manipulating articulated objects, performing long-horizon tasks, and demonstrating skills involving tool use and functional motion. Together, they provide a comprehensive evaluation across various task settings.

We also conduct extensive real-world experiments to prove the effectiveness of our data generation pipeline in terms of different types of generalization. 
Notably, the evaluation of object pose generalization is incorporated into all experiments, including those focused on the other five types of generalization (object types, camera views, embodiment types, lighting conditions, and scene appearance). 
This is because object pose generalization is a fundamental requirement for task completion ability.
For the other five types of generalization, the details are provided in Sec.~\ref{sec:aug_exp}.
Success rate (SR) is chosen as the evaluation metric in all experiments. 
Each policy is evaluated with 30 trials for a certain evaluation setting.

\subsection{Efficiency of Augmenting Demonstrations}
\label{sec:efficiency}

\begin{table*}[ht]
\centering
\caption{\textbf{Comparison of demonstration collection time (s).} We calculate the average time cost of data collection of a single demonstration over 100 demonstrations. Our method achieves more than 29 times the speed compared to the baseline.}
\begin{tabular}{c|ccccc|c}
\toprule
\textbf{Task Type} & \textbf{Pick Object} & \textbf{Close Drawer} & \textbf{Pick-Place-Close} & \textbf{Dual Pick-Place} & \textbf{Sweep} & \textbf{Average} \\
\midrule
 \textbf{Real-world} & 13.2 & 10.1 & 24.7 & 27.0 & 20.4 & 19.1 \\ 
 \rowcolor{cyan!10}
 \textbf{Ours} & \textbf{0.43} & \textbf{0.34} & \textbf{0.86} & \textbf{1.0} & \textbf{0.58} & \textbf{0.64} \\ 
\bottomrule
\end{tabular}
\label{tab:efficiency}
\end{table*}

To answer Question 1, we need to justify that our pipeline is economical with both labor and time when generating data.
The labor-saving property is obvious because demonstrations are generated automatically in our pipeline.
We compare the average time consumption of manually collecting a real-world demonstration to that of generating a demonstration through our pipeline. 
Specifically, we adopt eight processes on an NVIDIA RTX 4090 GPU for paralleled data generation to efficiently utilize computational resources.

The comparison study is conducted on all five tasks, and the result is shown in Table~\ref{tab:efficiency}. 
Our data generation pipeline that executed on a single GPU is more than 29 times faster than collecting data in the real world, with an average time consumption of 0.64s across all five tasks.
With no human interference, our demonstration generation approach is able to generate visually diverse training data with little time expenditure.

\begin{figure*}[ht]
    \centering
    \includegraphics[width=0.9\linewidth]{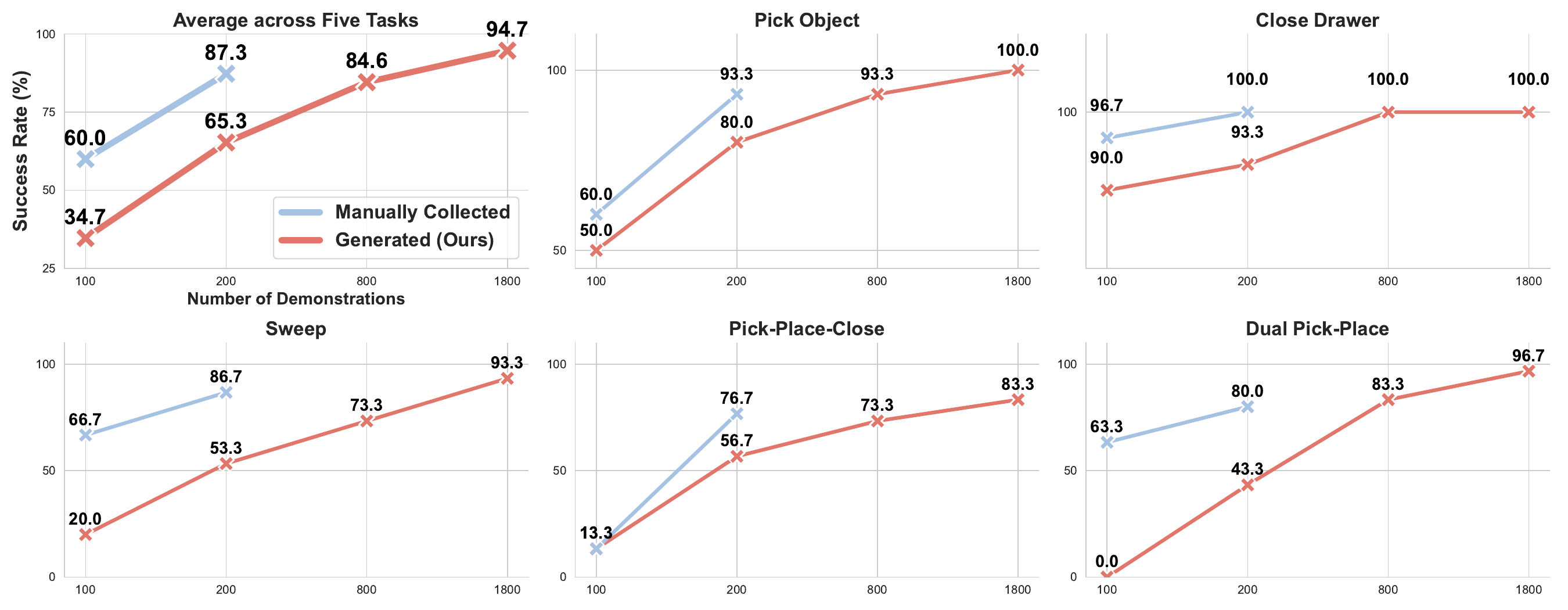}
    \vspace{0em}
    \caption{
    \textbf{Main results.} 
    \textbf{Top left:} We present the average success rate across five tasks. Our method shows promising scalability as the number of demonstration grows.
    \textbf{The other five subfigures:} For each task, we evaluate the success rate of policies trained from manually collected data and those generated by our method over 30 trials, using different number of demonstrations. 
    } 
    \label{fig:main_exp}
    \vspace{-3mm}
\end{figure*}

\subsection{Performance of the Policy Trained on Augmented Data}
\label{sec:main_exp}

To answer Question 2 and 3, we compare the policies trained on generated demonstrations and manually collected demonstrations in terms of their success rates when facing various object poses.
Moreover, we explore the performance of policies as generated data gradually scale up. 

The main results of the experiment are illustrated in Fig.~\ref{fig:main_exp}.
While policies trained on real-world demonstrations still have an edge over those trained on the same number of generated ones, our method manifests salient improvement in success rate as the generated demonstrations scale up. 
Concretely, visuomotor policies trained on 800 generated demonstrations achieve comparable performance to those trained on 200 manually collected demonstrations. Moreover, training with 1800 generated demonstrations raises the success rate to an average of 94.7\%, significantly surpassing the success rate achieved with 200 manually collected demonstrations.
It is also worth mentioning that the policy achieves a 96.7\% success rate on \textit{Dual Pick-Place} task with our generated data, which is nearly 20\% higher than the baseline (manually collected).
These findings testify the effectiveness of our method in generating novel object poses for better generalization of visuomotor policies, and indicate promising scaling property as generated data scales up.

\begin{figure*}[h]
    \centering
    \includegraphics[width=0.95\linewidth]{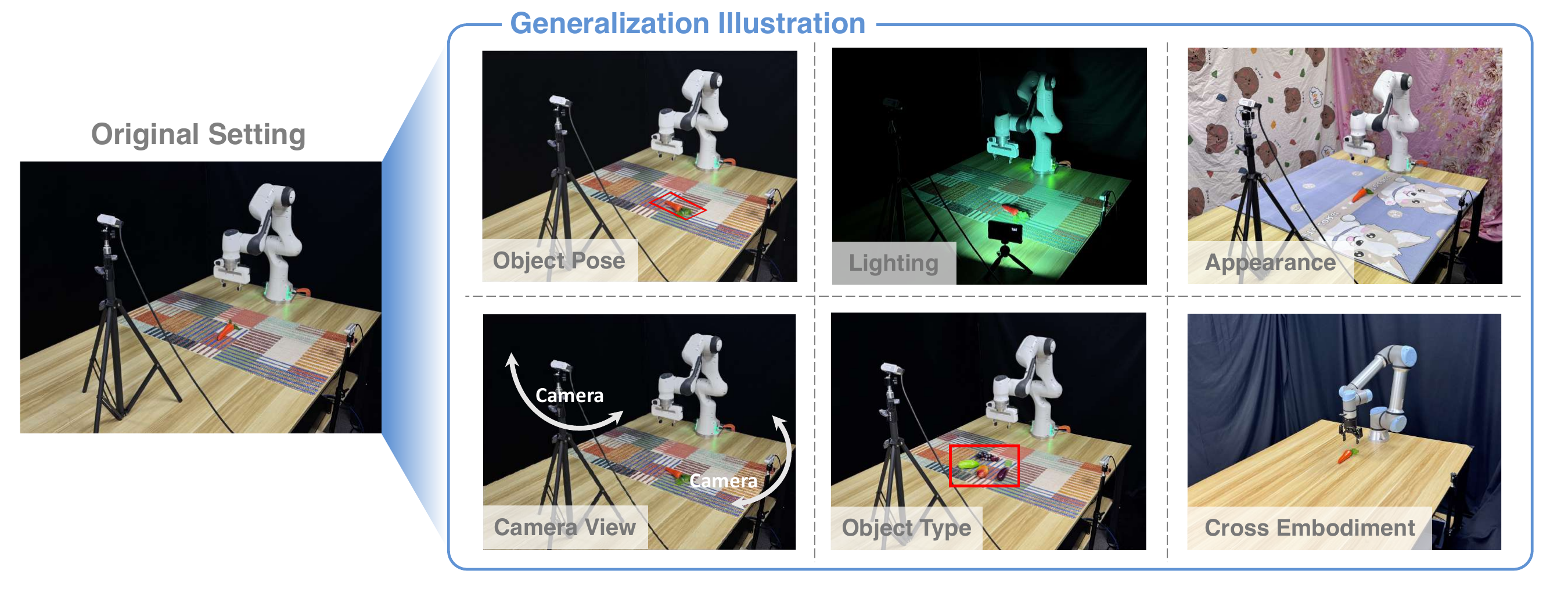}
    \vspace{-3mm}
    \caption{\textbf{Illustration of real-world experiments for different generalization types.} The data is collected in the original setting. When deploying the trained policy, we modify object poses, lighting conditions, scene appearance, camera views, object types, and embodiments to evaluate the robustness in different scenarios.}
    \label{fig:gen_illustration}
\end{figure*}

\subsection{Robustness when Facing Various Deployment Settings}
\label{sec:aug_exp}

To answer Question 4, we augment the expert demonstration in five different dimensions: lighting conditions, scene appearance, camera views, object types, and embodiment types. 
We compare policies trained on real-world data, real-world data augmented using 2D augmentation approaches, and data generated via our pipeline.
An illustration of the experiments for different generalization types is shown in Fig.~\ref{fig:gen_illustration}.

\begin{figure*}[ht]
    \centering
    \includegraphics[width=0.95\linewidth]{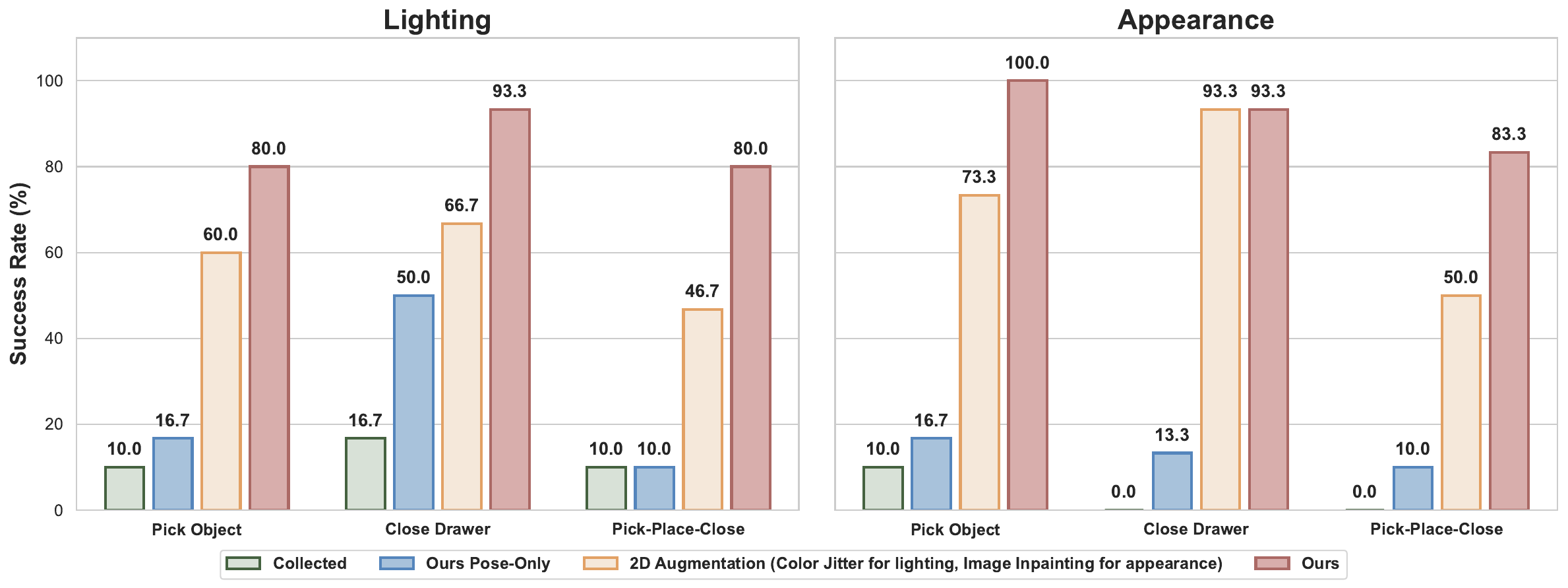}
    \vspace{0em}
    \caption{\textbf{Performance when changing lighting conditions and appearance.} We report the success rate of different policies under various lighting conditions and appearance. The policies trained with generated demonstrations with corresponding augmentations manifest remarkable advance compared to baseline policies.}
    \label{fig:lighting_appearance_exp}
    \vspace{-4mm}
\end{figure*}

\subsubsection{Lighting Condition}
\label{sec:lighting}

To demonstrate the effectiveness of lighting augmentation in our approach, we adopt five different scenarios for policy deployment, which are shown in Appendix~\ref{subsec:appearance_lighting_exp}.
We compare the performance of four policies that are trained respectively on:

\begin{enumerate}
    \item 200 real-world demonstrations (\textbf{Collected});
    \item 1800 generated demonstrations with only object pose augmentation, which are the same as data used in \ref{sec:main_exp} (\textbf{Ours Pose-Only});
    \item real-world demonstrations augmented with color jitter (\textbf{Color Jitter});
    \item 3200 demonstrations generated by our pipeline with both lighting condition and object pose augmentation (\textbf{Ours}).
\end{enumerate}

As shown in Fig.~\ref{fig:lighting_appearance_exp}, policies trained on augmented lighting conditions achieve an average of over 80\% success rate across \textit{Pick Object}, \textit{Close Drawer}, and \textit{Pick-Place-Close} tasks, with an overall improvement over those trained on real-world data without augmentation by 70\%.
Furthermore, our policies show a significant edge over those trained on generated demonstrations with augmented object poses and real-world demonstrations augmented with color jitter, justifying the validity of lighting augmentation in our pipeline.

\subsubsection{Scene Appearance}
\label{sec:appearance}

Similar to the experiment on lighting conditions, we select five different scenarios for evaluation on scene appearance augmentation, which is illustrated in Appendix~\ref{subsec:appearance_lighting_exp}.
The four policies for comparison are trained in a similar manner as described in Sec.~\ref{sec:lighting}, with the key difference being that we employ image inpainting methods ~\cite{yuan2025roboengine, chen2023genaug, yu2023scalingrobotlearningsemantically, chen2024semanticallycontrollableaugmentationsgeneralizable} as more robust and suitable 2D augmentation baselines for appearance generalization.
The results are shown in Fig.~\ref{fig:lighting_appearance_exp}.
The policy trained on data generated through our pipeline, incorporating both appearance and object pose augmentations, achieves superior performance compared to all baselines. Notably, it demonstrates over a 70\% increase in success rates across all three tasks when compared to policies trained on data without appearance augmentation.
In particular, our policy achieves 100\% success rate on the \textit{Pick Object} task, showcasing strong robustness against various background appearance.

\begin{table*}[ht]
\centering
\caption{\textbf{Performance when changing camera view.} We compare the success rate of different policies under two circumstances: novel camera view and moving camera view. The policies trained on demonstrations augmented using our approach showcase significant improvement over baseline policies.}


\resizebox{\textwidth}{!}{
\begin{tabular}{c|cc|cc|cc|c}
\toprule
\multirow{2}{*}{\textbf{Data Source}} & \multicolumn{2}{|c|}{\textbf{Pick Object}} & \multicolumn{2}{c|}{\textbf{Close Drawer}} & \multicolumn{2}{c|}{\textbf{Pick-Place-Close}} & \multirow{2}{*}{\textbf{Average}} \\ 
\cmidrule(lr){2-3} \cmidrule(lr){4-5} \cmidrule(lr){6-7}
 & Novel View & Moving View & Novel View & Moving View & Novel View & Moving View & \\ 
\midrule
 \textbf{Collected} & 6.7 & 0.0  & 16.7 & 13.3 & 0.0 & 0.0 & 6.1 \\ 
 \textbf{Ours Pose-Only} & 0.0 & 0.0 & 26.7 & 30.0 & 0.0 & 0.0 & 9.5 \\
 \textbf{VISTA~\cite{tian2024view}} & 33.3 & 33.3 & 56.7 & 70.0 & 33.3 & 16.7 & 40.6 \\
 \rowcolor{cyan!10}
 \textbf{Ours} & \textbf{90.0} & \textbf{86.7} & \textbf{100.0} & \textbf{96.7} & \textbf{53.3} & \textbf{56.7} & \textbf{80.6}\\ 
\bottomrule
\end{tabular}}

\label{tab:camera}
\end{table*}

\subsubsection{Camera View}
\label{sec:camera_view}

We employ two different settings for camera view generalization: \textit{novel view} and \textit{moving view}.
In \textit{novel view} experiments, we select 30 poses for each camera, which are different from the training perspective.
On the other hand, cameras are kept moving in \textit{moving view} experiments. 
Similar to Sec.~\ref{sec:lighting} and Sec.~\ref{sec:appearance}, we compare the performance of four policies that are trained respectively on:

\begin{enumerate}
\item 200 real-world demonstrations (\textbf{Collected});
\item 1800 generated demonstrations with only object pose augmentation (\textbf{Ours Pose-Only});
\item 3200 demonstrations stemmed from 200 real-world demonstrations, augmented using \textbf{VISTA}~\cite{tian2024view}, which leverages novel view synthesis models to augment data from different views;
\item 3200 generated demonstrations with camera view augmentation (\textbf{Ours}).
\end{enumerate}

We present the results in Table~\ref{tab:camera}.
Our policy is able to perform \textit{Pick Object} task and \textit{Pick-Place-Close} task with success rates of over 80\% and 50\% respectively, while the policies trained on data without augmentation can barely accomplish the task. 
Our approach also outperforms VISTA by a large margin.
Notably, our policy achieves nearly 100\% success rate on \textit{Close Drawer} task, manifesting strong robustness against novel camera views and moving cameras.

\subsubsection{Object Type}
\label{sec:object_type}

In order to demonstrate the effectiveness of our method in augmenting object types, we compare the performance of three different policies that are respectively trained on:

\begin{enumerate}
    \item 400 real-world demonstrations with 5 real-world objects (\textbf{Collected});
    \item 6400 demonstrations stemmed from 200 real-world demonstrations, augmented using \textbf{ROSIE}~\cite{yu2023scalingrobotlearningsemantically}, which utilizes image inpainting models to generate data with unseen objects;
    \item 6400 demonstrations generated by our pipeline with object type augmentation (\textbf{Ours}).
\end{enumerate}

\begin{figure}[h!]
    \centering
    \includegraphics[width=0.95\linewidth]{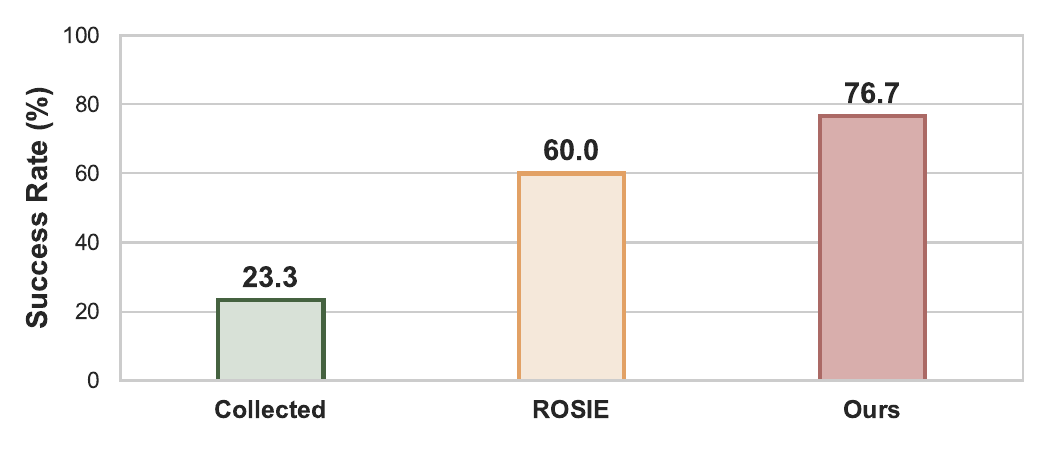}
    \vspace{-4mm}
    \caption{\textbf{Performance on novel object types.} The policy trained on data generated by RoboSplat shows a salient edge over baseline policies.}
    \label{fig:object_type}
    \vspace{-6mm}
\end{figure}

During deployment, we select five real-word objects that are different from all the objects covered in training process.
We report the result in Fig.~\ref{fig:object_type}.
The policy trained on 50 object types showcases better adaptability to novel object types, improving the success rate of baseline models by over 40\%.
This demonstrates the effectiveness of our data generation pipeline in utilizing off-the-shelf 3D Content Generation models to generalize policy to novel objects.

\subsubsection{Embodiment Type}
\label{sec:embodiment_type}

Our method supports generating demonstrations across different embodiment types as mentioned in Sec.~\ref{subsec:embodiment}. 
To prove that, based on one demonstration collected with the Franka Research 3, we generate novel demonstrations for a UR5e robot equipped with a Robotiq 2F-85 gripper and deploy the learned policy directly in the real world.
It is worth noting that policies trained on Franka Research 3 robot demonstrations fail to be deployed on UR5e robot due to frequent safety violations.
We compare the performance of policies trained on embodiment-augmented demonstrations with those trained on data augmented using RoVi-Aug~\cite{chen2024roviaugrobotviewpointaugmentation}. RoVi-Aug modifies real-world demonstrations by replacing the appearance of the embodiment through generative models.

\begin{figure}[h!]
    \centering
    \includegraphics[width=\linewidth]{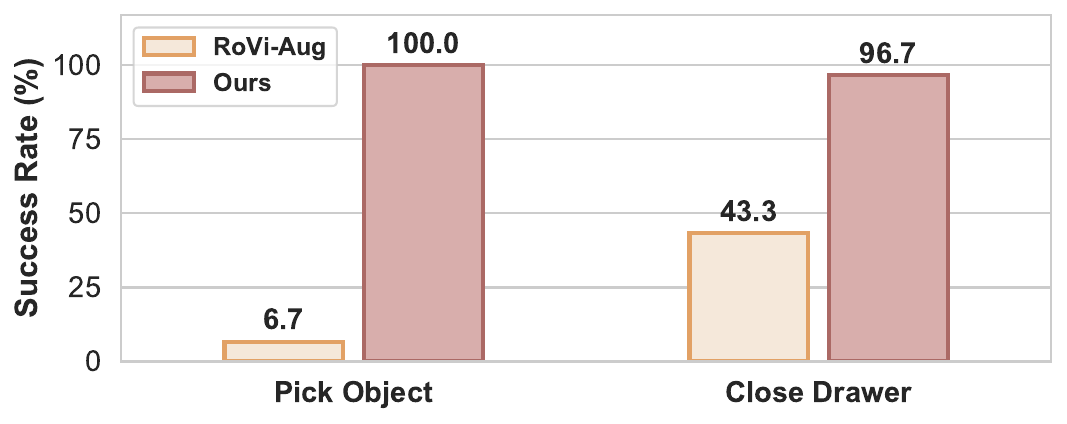}
    \caption{\textbf{Performance on cross embodiment experiments.} We evaluate the learned policy directly on the UR5e robot and achieve a nearly 100\% success rate that surpasses the 2D augmentation methods.}
    \label{fig:embodiment}
\end{figure}

We present the performance of policies in Fig.~\ref{fig:embodiment}.
Policies trained on data generated using our pipeline achieve a success rate close to 100\% on an embodiment different from the one used for demonstration collection. This result highlights its superior performance compared to the baseline in cross-embodiment transfer.



\section{Limitations}
\label{sec:limitations}

Due to the limitations of naive 3D Gaussian Splatting, it is incapable of handling deformable objects. Additionally, the pipeline lacks physical constraints, making it unsuitable for contact-rich and dynamic tasks. However, recent advancements in Gaussian Splatting~\cite{physgaussian, abouphysically, deformablegaussian, rong2024gaussian} provide promising opportunities to address these challenges. Future work could apply these techniques to generate data for a wider range of tasks.

\section{Conclusion} 
\label{sec:conclusion}

In this work, we introduce \textbf{RoboSplat}, a novel demonstration generation approach that requires only a single collected demonstration and generates diverse and high-quality data for policy learning. Comprehensive real-world experiments show that our approach significantly enhances the robustness of visuomotor policies when encountering various disturbances.

\section*{Acknowledgments}
We sincerely thank Yang Tian and Xiao Chen for their fruitful discussions.
This work is supported by the National Key R\&D Program of China (2022ZD0160201), Shanghai Artificial Intelligence Laboratory, and China Postdoctoral Science Foundation (2023M741848).


\bibliographystyle{plainnat}
\bibliography{references}

\begin{thebibliography}{71}
\providecommand{\natexlab}[1]{#1}
\providecommand{\url}[1]{\texttt{#1}}
\expandafter\ifx\csname urlstyle\endcsname\relax
  \providecommand{\doi}[1]{doi: #1}\else
  \providecommand{\doi}{doi: \begingroup \urlstyle{rm}\Url}\fi

\bibitem[Abou-Chakra et~al.(2024)Abou-Chakra, Rana, Dayoub, and Suenderhauf]{abouphysically}
Jad Abou-Chakra, Krishan Rana, Feras Dayoub, and Niko Suenderhauf.
\newblock Physically embodied gaussian splatting: A visually learnt and physically grounded 3d representation for robotics.
\newblock In \emph{8th Annual Conference on Robot Learning}, 2024.

\bibitem[Achiam et~al.(2023)Achiam, Adler, Agarwal, Ahmad, Akkaya, Aleman, Almeida, Altenschmidt, Altman, Anadkat, et~al.]{gpt4}
Josh Achiam, Steven Adler, Sandhini Agarwal, Lama Ahmad, Ilge Akkaya, Florencia~Leoni Aleman, Diogo Almeida, Janko Altenschmidt, Sam Altman, Shyamal Anadkat, et~al.
\newblock Gpt-4 technical report.
\newblock \emph{arXiv preprint arXiv:2303.08774}, 2023.

\bibitem[Ameperosa et~al.(2024)Ameperosa, Collins, Jain, and Garg]{ameperosa2024rocoda}
Ezra Ameperosa, Jeremy~A Collins, Mrinal Jain, and Animesh Garg.
\newblock Rocoda: Counterfactual data augmentation for data-efficient robot learning from demonstrations.
\newblock \emph{arXiv preprint arXiv:2411.16959}, 2024.

\bibitem[Barron et~al.(2022)Barron, Mildenhall, Verbin, Srinivasan, and Hedman]{barron2022mip}
Jonathan~T Barron, Ben Mildenhall, Dor Verbin, Pratul~P Srinivasan, and Peter Hedman.
\newblock Mip-nerf 360: Unbounded anti-aliased neural radiance fields.
\newblock In \emph{Proceedings of the IEEE/CVF conference on computer vision and pattern recognition}, pages 5470--5479, 2022.

\bibitem[Besl and McKay(1992)]{besl1992method}
Paul~J Besl and Neil~D McKay.
\newblock Method for registration of 3-d shapes.
\newblock In \emph{Sensor fusion IV: control paradigms and data structures}, volume 1611, pages 586--606. Spie, 1992.

\bibitem[Biza et~al.(2023)Biza, Thompson, Pagidi, Kumar, van~der Pol, Walters, Kipf, van~de Meent, Wong, and Platt]{biza2023oneshotimitationlearninginteraction}
Ondrej Biza, Skye Thompson, Kishore~Reddy Pagidi, Abhinav Kumar, Elise van~der Pol, Robin Walters, Thomas Kipf, Jan-Willem van~de Meent, Lawson~LS Wong, and Robert Platt.
\newblock One-shot imitation learning via interaction warping.
\newblock \emph{arXiv preprint arXiv:2306.12392}, 2023.

\bibitem[Brohan et~al.(2023)Brohan, Brown, Carbajal, Chebotar, Chen, Choromanski, Ding, Driess, Dubey, Finn, et~al.]{brohan2023rt2visionlanguageactionmodelstransfer}
Anthony Brohan, Noah Brown, Justice Carbajal, Yevgen Chebotar, Xi~Chen, Krzysztof Choromanski, Tianli Ding, Danny Driess, Avinava Dubey, Chelsea Finn, et~al.
\newblock Rt-2: Vision-language-action models transfer web knowledge to robotic control.
\newblock \emph{arXiv preprint arXiv:2307.15818}, 2023.

\bibitem[Chen et~al.(2024{\natexlab{a}})Chen, Xu, Dharmarajan, Irshad, Cheng, Keutzer, Tomizuka, Vuong, and Goldberg]{chen2024roviaugrobotviewpointaugmentation}
Lawrence~Yunliang Chen, Chenfeng Xu, Karthik Dharmarajan, Muhammad~Zubair Irshad, Richard Cheng, Kurt Keutzer, Masayoshi Tomizuka, Quan Vuong, and Ken Goldberg.
\newblock Rovi-aug: Robot and viewpoint augmentation for cross-embodiment robot learning.
\newblock \emph{arXiv preprint arXiv:2409.03403}, 2024{\natexlab{a}}.

\bibitem[Chen et~al.(2023)Chen, Kiami, Gupta, and Kumar]{chen2023genaug}
Zoey Chen, Sho Kiami, Abhishek Gupta, and Vikash Kumar.
\newblock Genaug: Retargeting behaviors to unseen situations via generative augmentation.
\newblock \emph{arXiv preprint arXiv:2302.06671}, 2023.

\bibitem[Chen et~al.(2024{\natexlab{b}})Chen, Mandi, Bharadhwaj, Sharma, Song, Gupta, and Kumar]{chen2024semanticallycontrollableaugmentationsgeneralizable}
Zoey Chen, Zhao Mandi, Homanga Bharadhwaj, Mohit Sharma, Shuran Song, Abhishek Gupta, and Vikash Kumar.
\newblock Semantically controllable augmentations for generalizable robot learning.
\newblock \emph{The International Journal of Robotics Research}, page 02783649241273686, 2024{\natexlab{b}}.

\bibitem[Chi et~al.(2023)Chi, Xu, Feng, Cousineau, Du, Burchfiel, Tedrake, and Song]{chi2024diffusionpolicyvisuomotorpolicy}
Cheng Chi, Zhenjia Xu, Siyuan Feng, Eric Cousineau, Yilun Du, Benjamin Burchfiel, Russ Tedrake, and Shuran Song.
\newblock Diffusion policy: Visuomotor policy learning via action diffusion.
\newblock \emph{The International Journal of Robotics Research}, page 02783649241273668, 2023.

\bibitem[Chi et~al.(2024)Chi, Xu, Pan, Cousineau, Burchfiel, Feng, Tedrake, and Song]{chi2024universal}
Cheng Chi, Zhenjia Xu, Chuer Pan, Eric Cousineau, Benjamin Burchfiel, Siyuan Feng, Russ Tedrake, and Shuran Song.
\newblock Universal manipulation interface: In-the-wild robot teaching without in-the-wild robots.
\newblock \emph{arXiv preprint arXiv:2402.10329}, 2024.

\bibitem[Chun et~al.(2023)Chun, Du, Simeonov, Lozano-Perez, and Kaelbling]{chun2023localneuraldescriptorfields}
Ethan Chun, Yilun Du, Anthony Simeonov, Tomas Lozano-Perez, and Leslie Kaelbling.
\newblock Local neural descriptor fields: Locally conditioned object representations for manipulation.
\newblock In \emph{2023 IEEE International Conference on Robotics and Automation (ICRA)}, pages 1830--1836. IEEE, 2023.

\bibitem[Dalal et~al.(2024)Dalal, Liu, Talbott, Chen, Pathak, Zhang, and Salakhutdinov]{dalal2024local}
Murtaza Dalal, Min Liu, Walter Talbott, Chen Chen, Deepak Pathak, Jian Zhang, and Ruslan Salakhutdinov.
\newblock Local policies enable zero-shot long-horizon manipulation.
\newblock \emph{arXiv preprint arXiv:2410.22332}, 2024.

\bibitem[Fan et~al.(2021)Fan, Wang, Huang, Yu, Fei-Fei, Zhu, and Anandkumar]{fan2021secantselfexpertcloningzeroshot}
Linxi Fan, Guanzhi Wang, De-An Huang, Zhiding Yu, Li~Fei-Fei, Yuke Zhu, and Anima Anandkumar.
\newblock Secant: Self-expert cloning for zero-shot generalization of visual policies.
\newblock \emph{arXiv preprint arXiv:2106.09678}, 2021.

\bibitem[Fang et~al.(2023)Fang, Wang, Fang, Gou, Liu, Yan, Liu, Xie, and Lu]{fang2023anygrasp}
Hao-Shu Fang, Chenxi Wang, Hongjie Fang, Minghao Gou, Jirong Liu, Hengxu Yan, Wenhai Liu, Yichen Xie, and Cewu Lu.
\newblock Anygrasp: Robust and efficient grasp perception in spatial and temporal domains.
\newblock \emph{IEEE Transactions on Robotics}, 2023.

\bibitem[Gao et~al.(2025)Gao, Gu, Lin, Li, Zhu, Cao, Zhang, and Yao]{gao2024relightable3dgaussiansrealistic}
Jian Gao, Chun Gu, Youtian Lin, Zhihao Li, Hao Zhu, Xun Cao, Li~Zhang, and Yao Yao.
\newblock Relightable 3d gaussians: Realistic point cloud relighting with brdf decomposition and ray tracing.
\newblock In \emph{European Conference on Computer Vision}, pages 73--89. Springer, 2025.

\bibitem[Haldar et~al.(2024)Haldar, Peng, and Pinto]{haldar2024baku}
Siddhant Haldar, Zhuoran Peng, and Lerrel Pinto.
\newblock Baku: An efficient transformer for multi-task policy learning.
\newblock \emph{arXiv preprint arXiv:2406.07539}, 2024.

\bibitem[Hansen and Wang(2021)]{hansen2021generalizationreinforcementlearningsoft}
Nicklas Hansen and Xiaolong Wang.
\newblock Generalization in reinforcement learning by soft data augmentation.
\newblock In \emph{2021 IEEE International Conference on Robotics and Automation (ICRA)}, pages 13611--13617. IEEE, 2021.

\bibitem[Hansen et~al.(2021)Hansen, Su, and Wang]{hansen2021stabilizingdeepqlearningconvnets}
Nicklas Hansen, Hao Su, and Xiaolong Wang.
\newblock Stabilizing deep q-learning with convnets and vision transformers under data augmentation.
\newblock \emph{Advances in neural information processing systems}, 34:\penalty0 3680--3693, 2021.

\bibitem[He et~al.(2016)He, Zhang, Ren, and Sun]{resnet}
Kaiming He, Xiangyu Zhang, Shaoqing Ren, and Jian Sun.
\newblock Deep residual learning for image recognition.
\newblock In \emph{Proceedings of the IEEE conference on computer vision and pattern recognition}, pages 770--778, 2016.

\bibitem[Hedman et~al.(2018)Hedman, Philip, Price, Frahm, Drettakis, and Brostow]{hedman2018deep}
Peter Hedman, Julien Philip, True Price, Jan-Michael Frahm, George Drettakis, and Gabriel Brostow.
\newblock Deep blending for free-viewpoint image-based rendering.
\newblock \emph{ACM Transactions on Graphics (ToG)}, 37\penalty0 (6):\penalty0 1--15, 2018.

\bibitem[Irpan et~al.(2022)Irpan, Herzog, Toshev, Zeng, Brohan, Ichter, David, Parada, Finn, Tan, et~al.]{irpan2022can}
Alex Irpan, Alexander Herzog, Alexander~Toshkov Toshev, Andy Zeng, Anthony Brohan, Brian~Andrew Ichter, Byron David, Carolina Parada, Chelsea Finn, Clayton Tan, et~al.
\newblock Do as i can, not as i say: Grounding language in robotic affordances.
\newblock In \emph{Conference on Robot Learning}, number 2022, 2022.

\bibitem[Ji et~al.(2024)Ji, Qiu, Zou, and Wang]{ji2024graspsplatsefficientmanipulation3d}
Mazeyu Ji, Ri-Zhao Qiu, Xueyan Zou, and Xiaolong Wang.
\newblock Graspsplats: Efficient manipulation with 3d feature splatting.
\newblock \emph{arXiv preprint arXiv:2409.02084}, 2024.

\bibitem[Kerbl et~al.(2023)Kerbl, Kopanas, Leimk{\"u}hler, and Drettakis]{kerbl20233dgaussiansplattingrealtime}
Bernhard Kerbl, Georgios Kopanas, Thomas Leimk{\"u}hler, and George Drettakis.
\newblock 3d gaussian splatting for real-time radiance field rendering.
\newblock \emph{ACM Trans. Graph.}, 42\penalty0 (4):\penalty0 139--1, 2023.

\bibitem[Knapitsch et~al.(2017)Knapitsch, Park, Zhou, and Koltun]{Knapitsch2017}
Arno Knapitsch, Jaesik Park, Qian-Yi Zhou, and Vladlen Koltun.
\newblock Tanks and temples: Benchmarking large-scale scene reconstruction.
\newblock \emph{ACM Transactions on Graphics}, 36\penalty0 (4), 2017.

\bibitem[Kopanas et~al.(2022)Kopanas, Leimk{\"u}hler, Rainer, Jambon, and Drettakis]{kopanas2022neural}
Georgios Kopanas, Thomas Leimk{\"u}hler, Gilles Rainer, Cl{\'e}ment Jambon, and George Drettakis.
\newblock Neural point catacaustics for novel-view synthesis of reflections.
\newblock \emph{ACM Transactions on Graphics (TOG)}, 41\penalty0 (6):\penalty0 1--15, 2022.

\bibitem[Kostrikov et~al.(2020)Kostrikov, Yarats, and Fergus]{kostrikov2021imageaugmentationneedregularizing}
Ilya Kostrikov, Denis Yarats, and Rob Fergus.
\newblock Image augmentation is all you need: Regularizing deep reinforcement learning from pixels.
\newblock \emph{arXiv preprint arXiv:2004.13649}, 2020.

\bibitem[Laskin et~al.(2020)Laskin, Lee, Stooke, Pinto, Abbeel, and Srinivas]{laskin2020reinforcementlearningaugmenteddata}
Misha Laskin, Kimin Lee, Adam Stooke, Lerrel Pinto, Pieter Abbeel, and Aravind Srinivas.
\newblock Reinforcement learning with augmented data.
\newblock \emph{Advances in neural information processing systems}, 33:\penalty0 19884--19895, 2020.

\bibitem[Levy et~al.(2024)Levy, Haldar, Pinto, and Shirivastava]{levy2024p3}
Mara Levy, Siddhant Haldar, Lerrel Pinto, and Abhinav Shirivastava.
\newblock P3-po: Prescriptive point priors for visuo-spatial generalization of robot policies.
\newblock \emph{arXiv preprint arXiv:2412.06784}, 2024.

\bibitem[Li et~al.(2024)Li, Li, Zhang, Zhang, Jia, Wang, Fan, Tseng, and Wang]{li2024robogsimreal2sim2realroboticgaussian}
Xinhai Li, Jialin Li, Ziheng Zhang, Rui Zhang, Fan Jia, Tiancai Wang, Haoqiang Fan, Kuo-Kun Tseng, and Ruiping Wang.
\newblock Robogsim: A real2sim2real robotic gaussian splatting simulator.
\newblock \emph{arXiv preprint arXiv:2411.11839}, 2024.

\bibitem[Liang et~al.(2024)Liang, Zhang, Feng, Shan, and Jia]{liang2024gsir3dgaussiansplatting}
Zhihao Liang, Qi~Zhang, Ying Feng, Ying Shan, and Kui Jia.
\newblock Gs-ir: 3d gaussian splatting for inverse rendering.
\newblock In \emph{Proceedings of the IEEE/CVF Conference on Computer Vision and Pattern Recognition}, pages 21644--21653, 2024.

\bibitem[Lin et~al.(2014)Lin, Maire, Belongie, Hays, Perona, Ramanan, Doll{\'a}r, and Zitnick]{lin2014microsoft}
Tsung-Yi Lin, Michael Maire, Serge Belongie, James Hays, Pietro Perona, Deva Ramanan, Piotr Doll{\'a}r, and C~Lawrence Zitnick.
\newblock Microsoft coco: Common objects in context.
\newblock In \emph{Computer Vision--ECCV 2014: 13th European Conference, Zurich, Switzerland, September 6-12, 2014, Proceedings, Part V 13}, pages 740--755. Springer, 2014.

\bibitem[Lu et~al.(2025)Lu, Zhang, Wang, Liu, Lu, and Tang]{lu2024manigaussiandynamicgaussiansplatting}
Guanxing Lu, Shiyi Zhang, Ziwei Wang, Changliu Liu, Jiwen Lu, and Yansong Tang.
\newblock Manigaussian: Dynamic gaussian splatting for multi-task robotic manipulation.
\newblock In \emph{European Conference on Computer Vision}, pages 349--366. Springer, 2025.

\bibitem[Mandi et~al.(2022)Mandi, Bharadhwaj, Moens, Song, Rajeswaran, and Kumar]{mandi2023cactiframeworkscalablemultitask}
Zhao Mandi, Homanga Bharadhwaj, Vincent Moens, Shuran Song, Aravind Rajeswaran, and Vikash Kumar.
\newblock Cacti: A framework for scalable multi-task multi-scene visual imitation learning.
\newblock \emph{arXiv preprint arXiv:2212.05711}, 2022.

\bibitem[Mandlekar et~al.(2021)Mandlekar, Xu, Wong, Nasiriany, Wang, Kulkarni, Fei-Fei, Savarese, Zhu, and Mart{\'\i}n-Mart{\'\i}n]{mandlekar2021matterslearningofflinehuman}
Ajay Mandlekar, Danfei Xu, Josiah Wong, Soroush Nasiriany, Chen Wang, Rohun Kulkarni, Li~Fei-Fei, Silvio Savarese, Yuke Zhu, and Roberto Mart{\'\i}n-Mart{\'\i}n.
\newblock What matters in learning from offline human demonstrations for robot manipulation.
\newblock \emph{arXiv preprint arXiv:2108.03298}, 2021.

\bibitem[Mandlekar et~al.(2023)Mandlekar, Nasiriany, Wen, Akinola, Narang, Fan, Zhu, and Fox]{mandlekar2023mimicgendatagenerationscalable}
Ajay Mandlekar, Soroush Nasiriany, Bowen Wen, Iretiayo Akinola, Yashraj Narang, Linxi Fan, Yuke Zhu, and Dieter Fox.
\newblock Mimicgen: A data generation system for scalable robot learning using human demonstrations.
\newblock \emph{arXiv preprint arXiv:2310.17596}, 2023.

\bibitem[{Octo Model Team} et~al.(2024){Octo Model Team}, Ghosh, Walke, Pertsch, Black, Mees, Dasari, Hejna, Xu, Luo, Kreiman, Tan, Chen, Sanketi, Vuong, Xiao, Sadigh, Finn, and Levine]{octo}
{Octo Model Team}, Dibya Ghosh, Homer Walke, Karl Pertsch, Kevin Black, Oier Mees, Sudeep Dasari, Joey Hejna, Charles Xu, Jianlan Luo, Tobias Kreiman, {You Liang} Tan, Lawrence~Yunliang Chen, Pannag Sanketi, Quan Vuong, Ted Xiao, Dorsa Sadigh, Chelsea Finn, and Sergey Levine.
\newblock Octo: An open-source generalist robot policy.
\newblock In \emph{Proceedings of Robotics: Science and Systems}, Delft, Netherlands, 2024.

\bibitem[O'Neill et~al.(2023)O'Neill, Rehman, Gupta, Maddukuri, Gupta, Padalkar, Lee, Pooley, Gupta, Mandlekar, et~al.]{embodimentcollaboration2024openxembodimentroboticlearning}
Abby O'Neill, Abdul Rehman, Abhinav Gupta, Abhiram Maddukuri, Abhishek Gupta, Abhishek Padalkar, Abraham Lee, Acorn Pooley, Agrim Gupta, Ajay Mandlekar, et~al.
\newblock Open x-embodiment: Robotic learning datasets and rt-x models.
\newblock \emph{arXiv preprint arXiv:2310.08864}, 2023.

\bibitem[Qureshi et~al.(2024)Qureshi, Garg, Yandun, Held, Kantor, and Silwal]{qureshi2024splatsimzeroshotsim2realtransfer}
Mohammad~Nomaan Qureshi, Sparsh Garg, Francisco Yandun, David Held, George Kantor, and Abhisesh Silwal.
\newblock Splatsim: Zero-shot sim2real transfer of rgb manipulation policies using gaussian splatting.
\newblock \emph{arXiv preprint arXiv:2409.10161}, 2024.

\bibitem[Ren et~al.(2024)Ren, Liu, Zeng, Lin, Li, Cao, Chen, Huang, Chen, Yan, et~al.]{ren2024grounded}
Tianhe Ren, Shilong Liu, Ailing Zeng, Jing Lin, Kunchang Li, He~Cao, Jiayu Chen, Xinyu Huang, Yukang Chen, Feng Yan, et~al.
\newblock Grounded sam: Assembling open-world models for diverse visual tasks.
\newblock \emph{arXiv preprint arXiv:2401.14159}, 2024.

\bibitem[Rong et~al.(2024)Rong, Grigorev, Wang, Black, Thomaszewski, Tsalicoglou, and Hilliges]{rong2024gaussian}
Boxiang Rong, Artur Grigorev, Wenbo Wang, Michael~J Black, Bernhard Thomaszewski, Christina Tsalicoglou, and Otmar Hilliges.
\newblock Gaussian garments: Reconstructing simulation-ready clothing with photorealistic appearance from multi-view video.
\newblock \emph{arXiv preprint arXiv:2409.08189}, 2024.

\bibitem[Ryu et~al.(2022)Ryu, Lee, Lee, and Choi]{ryu2023equivariantdescriptorfieldsse3equivariant}
Hyunwoo Ryu, Hong-in Lee, Jeong-Hoon Lee, and Jongeun Choi.
\newblock Equivariant descriptor fields: Se (3)-equivariant energy-based models for end-to-end visual robotic manipulation learning.
\newblock \emph{arXiv preprint arXiv:2206.08321}, 2022.

\bibitem[Sch\"{o}nberger and Frahm(2016)]{schoenberger2016sfm}
Johannes~Lutz Sch\"{o}nberger and Jan-Michael Frahm.
\newblock Structure-from-motion revisited.
\newblock In \emph{Conference on Computer Vision and Pattern Recognition (CVPR)}, 2016.

\bibitem[Sch\"{o}nberger et~al.(2016)Sch\"{o}nberger, Zheng, Pollefeys, and Frahm]{schoenberger2016mvs}
Johannes~Lutz Sch\"{o}nberger, Enliang Zheng, Marc Pollefeys, and Jan-Michael Frahm.
\newblock Pixelwise view selection for unstructured multi-view stereo.
\newblock In \emph{European Conference on Computer Vision (ECCV)}, 2016.

\bibitem[Seo et~al.(2023)Seo, Kim, James, Lee, Shin, and Abbeel]{mvmwm}
Younggyo Seo, Junsu Kim, Stephen James, Kimin Lee, Jinwoo Shin, and Pieter Abbeel.
\newblock Multi-view masked world models for visual robotic manipulation.
\newblock In \emph{International Conference on Machine Learning}, pages 30613--30632. PMLR, 2023.

\bibitem[Shorinwa et~al.(2024)Shorinwa, Tucker, Smith, Swann, Chen, Firoozi, Kennedy~III, and Schwager]{shorinwa2024splatmovermultistageopenvocabularyrobotic}
Ola Shorinwa, Johnathan Tucker, Aliyah Smith, Aiden Swann, Timothy Chen, Roya Firoozi, Monroe Kennedy~III, and Mac Schwager.
\newblock Splat-mover: Multi-stage, open-vocabulary robotic manipulation via editable gaussian splatting.
\newblock \emph{arXiv preprint arXiv:2405.04378}, 2024.

\bibitem[Simeonov et~al.(2022)Simeonov, Du, Tagliasacchi, Tenenbaum, Rodriguez, Agrawal, and Sitzmann]{simeonov2021neuraldescriptorfieldsse3equivariant}
Anthony Simeonov, Yilun Du, Andrea Tagliasacchi, Joshua~B Tenenbaum, Alberto Rodriguez, Pulkit Agrawal, and Vincent Sitzmann.
\newblock Neural descriptor fields: Se (3)-equivariant object representations for manipulation.
\newblock In \emph{2022 International Conference on Robotics and Automation (ICRA)}, pages 6394--6400. IEEE, 2022.

\bibitem[Singh et~al.(2024)Singh, Allshire, Handa, Ratliff, and Van~Wyk]{singh2024dextrah}
Ritvik Singh, Arthur Allshire, Ankur Handa, Nathan Ratliff, and Karl Van~Wyk.
\newblock Dextrah-rgb: Visuomotor policies to grasp anything with dexterous hands.
\newblock \emph{arXiv preprint arXiv:2412.01791}, 2024.

\bibitem[Tian et~al.(2024{\natexlab{a}})Tian, Wulfe, Sargent, Liu, Zakharov, Guizilini, and Wu]{tian2024view}
Stephen Tian, Blake Wulfe, Kyle Sargent, Katherine Liu, Sergey Zakharov, Vitor Guizilini, and Jiajun Wu.
\newblock View-invariant policy learning via zero-shot novel view synthesis.
\newblock \emph{arXiv preprint arXiv:2409.03685}, 2024{\natexlab{a}}.

\bibitem[Tian et~al.(2024{\natexlab{b}})Tian, Yang, Zeng, Wang, Lin, Dong, and Pang]{seer}
Yang Tian, Sizhe Yang, Jia Zeng, Ping Wang, Dahua Lin, Hao Dong, and Jiangmiao Pang.
\newblock Predictive inverse dynamics models are scalable learners for robotic manipulation.
\newblock \emph{arXiv preprint arXiv:2412.15109}, 2024{\natexlab{b}}.

\bibitem[Torne et~al.(2024)Torne, Simeonov, Li, Chan, Chen, Gupta, and Agrawal]{torne2024reconcilingrealitysimulationrealtosimtoreal}
Marcel Torne, Anthony Simeonov, Zechu Li, April Chan, Tao Chen, Abhishek Gupta, and Pulkit Agrawal.
\newblock Reconciling reality through simulation: A real-to-sim-to-real approach for robust manipulation.
\newblock \emph{arXiv preprint arXiv:2403.03949}, 2024.

\bibitem[Vitiello et~al.(2023)Vitiello, Dreczkowski, and Johns]{vitiello2023oneshotimitationlearningpose}
Pietro Vitiello, Kamil Dreczkowski, and Edward Johns.
\newblock One-shot imitation learning: A pose estimation perspective.
\newblock \emph{arXiv preprint arXiv:2310.12077}, 2023.

\bibitem[Vosylius and Johns(2024)]{vosylius2024instantpolicyincontextimitation}
Vitalis Vosylius and Edward Johns.
\newblock Instant policy: In-context imitation learning via graph diffusion.
\newblock \emph{arXiv preprint arXiv:2411.12633}, 2024.

\bibitem[Wu et~al.(2023)Wu, Jing, Cheang, Chen, Xu, Li, Liu, Li, and Kong]{gr1}
Hongtao Wu, Ya~Jing, Chilam Cheang, Guangzeng Chen, Jiafeng Xu, Xinghang Li, Minghuan Liu, Hang Li, and Tao Kong.
\newblock Unleashing large-scale video generative pre-training for visual robot manipulation, 2023.

\bibitem[Wu et~al.(2024)Wu, Pan, Wu, Wang, Miao, and Wang]{wu2024rlgsbridge3dgaussiansplatting}
Yuxuan Wu, Lei Pan, Wenhua Wu, Guangming Wang, Yanzi Miao, and Hesheng Wang.
\newblock Rl-gsbridge: 3d gaussian splatting based real2sim2real method for robotic manipulation learning.
\newblock \emph{arXiv preprint arXiv:2409.20291}, 2024.

\bibitem[Xiang et~al.(2024)Xiang, Lv, Xu, Deng, Wang, Zhang, Chen, Tong, and Yang]{xiang2024structured}
Jianfeng Xiang, Zelong Lv, Sicheng Xu, Yu~Deng, Ruicheng Wang, Bowen Zhang, Dong Chen, Xin Tong, and Jiaolong Yang.
\newblock Structured 3d latents for scalable and versatile 3d generation.
\newblock \emph{arXiv preprint arXiv:2412.01506}, 2024.

\bibitem[Xie et~al.(2024)Xie, Zong, Qiu, Li, Feng, Yang, and Jiang]{physgaussian}
Tianyi Xie, Zeshun Zong, Yuxing Qiu, Xuan Li, Yutao Feng, Yin Yang, and Chenfanfu Jiang.
\newblock Physgaussian: Physics-integrated 3d gaussians for generative dynamics.
\newblock In \emph{Proceedings of the IEEE/CVF Conference on Computer Vision and Pattern Recognition}, pages 4389--4398, 2024.

\bibitem[Xue et~al.(2025)Xue, Deng, Chen, Wang, Yuan, and Xu]{xue2025demogen}
Zhengrong Xue, Shuying Deng, Zhenyang Chen, Yixuan Wang, Zhecheng Yuan, and Huazhe Xu.
\newblock Demogen: Synthetic demonstration generation for data-efficient visuomotor policy learning.
\newblock \emph{arXiv preprint arXiv:2502.16932}, 2025.

\bibitem[Yang et~al.(2024{\natexlab{a}})Yang, Cao, Deng, Antonova, Song, and Bohg]{yang2024equibotsim3equivariantdiffusionpolicy}
Jingyun Yang, Zi-ang Cao, Congyue Deng, Rika Antonova, Shuran Song, and Jeannette Bohg.
\newblock Equibot: Sim (3)-equivariant diffusion policy for generalizable and data efficient learning.
\newblock \emph{arXiv preprint arXiv:2407.01479}, 2024{\natexlab{a}}.

\bibitem[Yang et~al.(2024{\natexlab{b}})Yang, Deng, Wu, Antonova, Guibas, and Bohg]{yang2024equivactsim3equivariantvisuomotorpolicies}
Jingyun Yang, Congyue Deng, Jimmy Wu, Rika Antonova, Leonidas Guibas, and Jeannette Bohg.
\newblock Equivact: Sim (3)-equivariant visuomotor policies beyond rigid object manipulation.
\newblock In \emph{2024 IEEE International Conference on Robotics and Automation (ICRA)}, pages 9249--9255. IEEE, 2024{\natexlab{b}}.

\bibitem[Yang et~al.(2024{\natexlab{c}})Yang, Kang, Huang, Xu, Feng, and Zhao]{yang2024depth}
Lihe Yang, Bingyi Kang, Zilong Huang, Xiaogang Xu, Jiashi Feng, and Hengshuang Zhao.
\newblock Depth anything: Unleashing the power of large-scale unlabeled data.
\newblock In \emph{Proceedings of the IEEE/CVF Conference on Computer Vision and Pattern Recognition}, pages 10371--10381, 2024{\natexlab{c}}.

\bibitem[Yang et~al.(2024{\natexlab{d}})Yang, Ze, and Xu]{movie}
Sizhe Yang, Yanjie Ze, and Huazhe Xu.
\newblock Movie: Visual model-based policy adaptation for view generalization.
\newblock \emph{Advances in Neural Information Processing Systems}, 36, 2024{\natexlab{d}}.

\bibitem[Yang et~al.(2024{\natexlab{e}})Yang, Gao, Zhou, Jiao, Zhang, and Jin]{deformablegaussian}
Ziyi Yang, Xinyu Gao, Wen Zhou, Shaohui Jiao, Yuqing Zhang, and Xiaogang Jin.
\newblock Deformable 3d gaussians for high-fidelity monocular dynamic scene reconstruction.
\newblock In \emph{Proceedings of the IEEE/CVF Conference on Computer Vision and Pattern Recognition}, pages 20331--20341, 2024{\natexlab{e}}.

\bibitem[Ye et~al.(2025)Ye, Danelljan, Yu, and Ke]{ye2024gaussiangroupingsegmentedit}
Mingqiao Ye, Martin Danelljan, Fisher Yu, and Lei Ke.
\newblock Gaussian grouping: Segment and edit anything in 3d scenes.
\newblock In \emph{European Conference on Computer Vision}, pages 162--179. Springer, 2025.

\bibitem[Yeshwanth et~al.(2023)Yeshwanth, Liu, Nie{\ss}ner, and Dai]{yeshwanth2023scannet++}
Chandan Yeshwanth, Yueh-Cheng Liu, Matthias Nie{\ss}ner, and Angela Dai.
\newblock Scannet++: A high-fidelity dataset of 3d indoor scenes.
\newblock In \emph{Proceedings of the IEEE/CVF International Conference on Computer Vision}, pages 12--22, 2023.

\bibitem[Yu et~al.(2023)Yu, Xiao, Stone, Tompson, Brohan, Wang, Singh, Tan, Peralta, Ichter, et~al.]{yu2023scalingrobotlearningsemantically}
Tianhe Yu, Ted Xiao, Austin Stone, Jonathan Tompson, Anthony Brohan, Su~Wang, Jaspiar Singh, Clayton Tan, Jodilyn Peralta, Brian Ichter, et~al.
\newblock Scaling robot learning with semantically imagined experience.
\newblock \emph{arXiv preprint arXiv:2302.11550}, 2023.

\bibitem[Yuan et~al.(2025)Yuan, Joshi, Zhu, Su, Zhao, and Gao]{yuan2025roboengine}
Chengbo Yuan, Suraj Joshi, Shaoting Zhu, Hang Su, Hang Zhao, and Yang Gao.
\newblock Roboengine: Plug-and-play robot data augmentation with semantic robot segmentation and background generation.
\newblock \emph{arXiv preprint arXiv:2503.18738}, 2025.

\bibitem[Yuan et~al.(2024)Yuan, Wei, Cheng, Zhang, Chen, and Xu]{yuan2024learningmanipulateanywherevisual}
Zhecheng Yuan, Tianming Wei, Shuiqi Cheng, Gu~Zhang, Yuanpei Chen, and Huazhe Xu.
\newblock Learning to manipulate anywhere: A visual generalizable framework for reinforcement learning.
\newblock \emph{arXiv preprint arXiv:2407.15815}, 2024.

\bibitem[Zhang and Boularias(2024)]{zhang2024oneshotimitationlearninginvariance}
Xinyu Zhang and Abdeslam Boularias.
\newblock One-shot imitation learning with invariance matching for robotic manipulation.
\newblock \emph{arXiv preprint arXiv:2405.13178}, 2024.

\bibitem[Zheng et~al.(2024)Zheng, Chen, Zheng, Gu, Yang, Jin, Li, Zhong, Wang, Liu, et~al.]{zheng2024gaussiangrasper3dlanguagegaussian}
Yuhang Zheng, Xiangyu Chen, Yupeng Zheng, Songen Gu, Runyi Yang, Bu~Jin, Pengfei Li, Chengliang Zhong, Zengmao Wang, Lina Liu, et~al.
\newblock Gaussiangrasper: 3d language gaussian splatting for open-vocabulary robotic grasping.
\newblock \emph{arXiv preprint arXiv:2403.09637}, 2024.

\end{thebibliography}

\clearpage

\appendix
\label{sec:appendix}
\subsection{Applying Transformation and Scaling to 3D Gaussians}
\label{subsec:applying_transformation}

This section outlines how to apply transformations (translation, rotation) and scaling to 3D Gaussians.

The Gaussian primitive typically possesses three core properties: 
1) a center position in three-dimensional space; 
2) an orientation that specifies the tilt of its principal axes, commonly represented as a quaternion; 
3) a scale indicating its width or narrowness. 
Additionally, Gaussian primitives can be enhanced with Spherical Harmonics (SH) to capture complex, direction-dependent color features.

When applying a transformation to the Gaussian primitive, the following steps should be taken: 
1) update the center position by scaling, rotating, and then adding the translation offset; 
2) update the orientation by combining the existing rotation with the new rotation; 
3) adjust the scale by multiplying by the scaling factor; 
4) rotate the Spherical Harmonics coefficients by using the Wigner D matrices. 

\subsection{Details of Demonstration Augmentation Process}
\label{subsec:aug_detail}

\begin{figure*}[ht]
    \centering
    \includegraphics[width=0.9\linewidth]{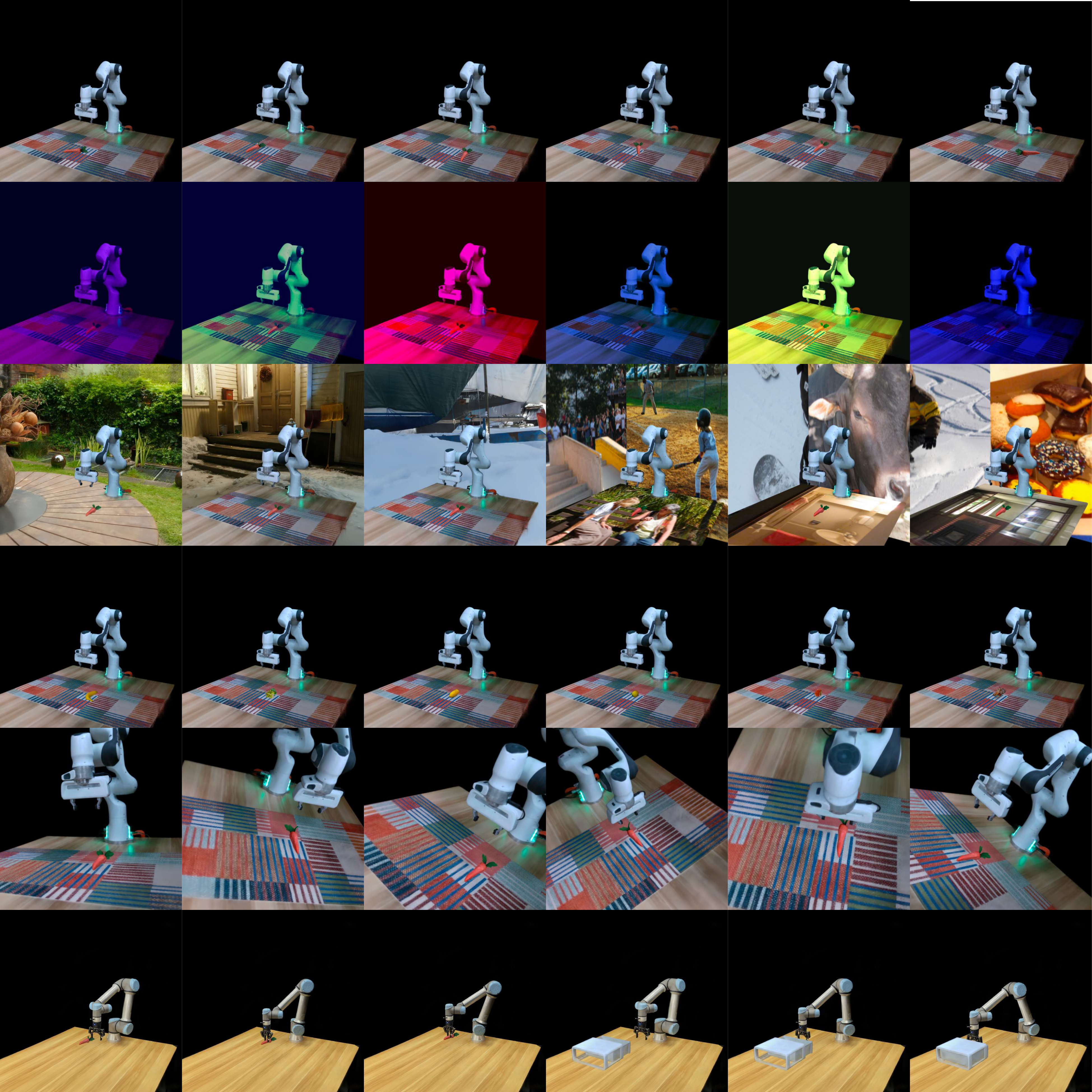}
    \vspace{0em}
    \caption{\textbf{Illustration of augmented demonstrations.} Type of generalization from the top row to the bottom row: object pose, lighting condition, scene appearance, object type, camera view, and embodiment type.}
    \label{fig:aug_illu}
    \vspace{-2mm}
\end{figure*}

We expand on the details of the demonstration augmentation process in this section.
An illustration of augmented demonstrations is provided in Fig.~\ref{fig:aug_illu}.

\subsubsection{Object pose}
\label{appendixsec:object_pose}

As mentioned in Sec.~\ref{subsec:object_pose}, we transform the end-effector poses at key frames equivariantly according to the transformation that is applied to the target object.
However, considering the symmetry of the gripper, we perform post-processing on the transformed end-effector pose.

Suppose the rotation of the transformed end-effector pose can be expressed as $(r_x, r_y, r_z)$ in the format of XYZ Euler angles.
We replace $r_z$ with $r_z'$, which can be calculated as:

\begin{align*}
    r_z'=
    \begin{cases}
        r_z & -\frac{\pi}{2}\leq r_z \leq \frac{\pi}{2} \\
        r_z + \pi & r_z < -\frac{\pi}{2}\\
        r_z - \pi & r_z > \frac{\pi}{2}.
    \end{cases}
    \label{eq:rot_sym}
\end{align*}

The resulting Euler angles $(r_x,r_y,r_z')$ form the final rotation of the end-effector, which prevents the end-effector from performing redundant rotation along its $z$-axis.

\subsubsection{Camera view}
\label{appendixsec:camera_view}

As aforementioned in Sec.~\ref{sec:camera_view}, we enumerate the hyperparameters of camera view augmentations and their range of randomization in Table~\ref{tab:camera_view_hyper}.
Suppose the camera view in the expert demonstration has target point $O_c^{\text{expert}}=(x_c^{0}, y_c^0, z_c^0)$ and corresponding spherical coordinates $(r^0,\theta^0,\varphi^0)$.
Thereby, the target point $O_c=(x_c, y_c, z_c)$ and corresponding spherical coordinates $(r,\theta,\varphi)$ are sampled from uniform distributions, ranging between $(x_c^0\pm\Delta x_c, y_c^0\pm\Delta y_c, z_c^0\pm\Delta z_c, r^0\pm\Delta r, \theta^0\pm\Delta \theta, \varphi^0\pm\Delta \varphi)$.

\begin{table}[h!]
\centering
\caption{\textbf{Camera view augmentation hyperparameters and their range of randomization.}}
\label{tab:camera_view_hyper}




\begin{tabular}{cc}
\toprule[1.5pt]
Hyperparameter & Value \\
\midrule[1.5pt]
$\Delta x_c$ & 0.1(m) \\
$\Delta y_c$ & 0.1(m) \\
$\Delta z_c$ & 0.1(m) \\
$\Delta r$ & 0.2(m) \\
$\Delta \theta$ & $\frac{\pi}{6}$ \\
$\Delta \varphi$ & $\frac{\pi}{6}$ \\
\bottomrule[1.5pt]
\end{tabular}

\end{table}

\subsubsection{Lighting condition}
\label{appendixsec:lighting}

We present the hyperparameters of lighting condition augmentation in this section. 
First, we normalize the RGB values of each pixel with minimum value $0$ and maximum value $1$. 
Then, we stipulate that the hyperparameters are sampled from the following distributions: 

\begin{align*}
    (\Delta_r, \Delta_g, \Delta_b) &\sim \mathcal{N}(\mathbf{0}, 0.1^2\mathbf{I}),\\
    s_r,s_g,s_b &\sim \text{Uniform}(0.3,1.8),\\
    o_r, o_g, o_b &\sim \text{Uniform}(-0.3, 0.3).
\end{align*}

\subsection{Policy Architecture}
\label{subsec:policy_architecture}

\begin{figure}[ht]
    \centering
    \vspace{2mm}
    \includegraphics[width=0.9\linewidth]{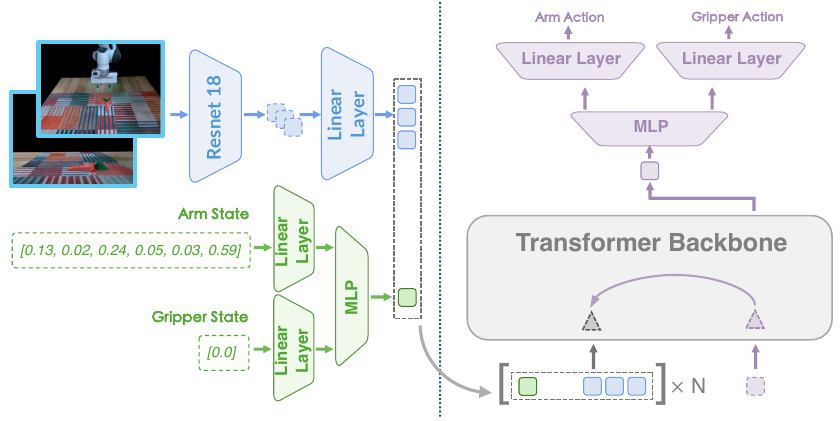}
    \vspace{0em}
    \caption{\textbf{Policy architecture.}}
    \label{fig:policy_architecture}
\end{figure}

As illustrated in Fig.~\ref{fig:policy_architecture}, the policy processes two types of inputs: images and robot states. 
We use different encoders to tokenize each modality accordingly.
For image inputs, the images are first passed through a ResNet-18 vision encoder to generate visual embeddings. We employ a linear layer to extract compact visual features.
For the robot state, we encode it into state tokens using a multi-layer perceptron (MLP).

The multi-modal encoder in our model is based on a GPT-2 style transformer architecture.
Before feeding the sequential image and state tokens into the transformer, we append readout tokens $[\text{ACT}]$ to the end. 
These readout tokens attend to embeddings from different modalities, serving as action latents used for action prediction.

Encoded by the multi-modal encoder, the action latents generated by the $[\text{ACT}]$ tokens are fed into the readout decoders to predict actions.
The action decoder utilizes an MLP to transform the action latent into the action vector.
We predict a chunk of 10 future actions. 
Compared to single-step action prediction, predicting multiple steps provides temporal action consistency and robustness to idle actions~\cite{chi2024diffusionpolicyvisuomotorpolicy}.

\subsection{Training Details}
\label{subsec:training_details}

During training, the input at each timestep consists of two images captured from two eye-on-base cameras, along with the robot state. The robot state includes both the arm state and the gripper state. The gripper state is binary, indicating whether the gripper is open or closed. For the Franka FR3 robot, the arm state is 7-dimensional, while for the UR5e robot, it is 6-dimensional.

The policy operates with a history length of 1, and the size of the action chunk is set to 10. During inference, we utilize temporal ensemble techniques to compute a weighted average of the multi-step actions.

The policy is trained using a single NVIDIA RTX 4090 GPU, with a batch size of 256 and a learning rate of 1e-4. Depending on the number of demonstrations, the policy is trained for varying numbers of epochs. The hyperparameters used during training are detailed in Table~\ref{tab:hyperparameters}.

\begin{table*}[ht]
\centering
\caption{
\textbf{Policy training hyperparameters.}
} \label{tab:hyperparameters}
\begin{tabular}{cc}
\toprule[1.5pt]

\multirow{2}{*}{Batch Size} & \multirow{2}{*}{256} \\ 
& \\\midrule
\multirow{2}{*}{Learning Rate} & \multirow{2}{*}{1e-4} \\
& \\\midrule

 & \multirow{2}{*}{1400 (100 demonstrations)} \\

 & \multirow{2}{*}{1000 (200 demonstrations)} \\

 & \multirow{2}{*}{800 (400 demonstrations)} \\

\multirow{2}{*}{Training Epochs} & \multirow{2}{*}{700 (800 demonstrations)} \\

 & \multirow{2}{*}{500 (1800 demonstrations)} \\

 & \multirow{2}{*}{300 (3200 demonstrations)} \\

 & \multirow{2}{*}{200 (6400 demonstrations)} \\

& \\\midrule
\multirow{2}{*}{Image Size} & \multirow{2}{*}{128*128} \\ 
& \\\midrule
\multirow{2}{*}{Optimizer} & \multirow{2}{*}{AdamW} \\
& \\ \midrule
\multirow{2}{*}{History Length} & \multirow{2}{*}{1} \\
& \\ \midrule
\multirow{2}{*}{\makecell[c]{Action Chunk\\Length}} & \multirow{2}{*}{10} \\
& \\

\bottomrule[1.5pt]

\end{tabular}

\end{table*}

\subsection{Illustration of Real-World Experiment Settings}
\label{subsec:appearance_lighting_exp}

\begin{figure*}[ht]
    \centering
    \subfigure[Flashing light (Red)]{\label{fig:red_flash}\includegraphics[width=0.3\textwidth]{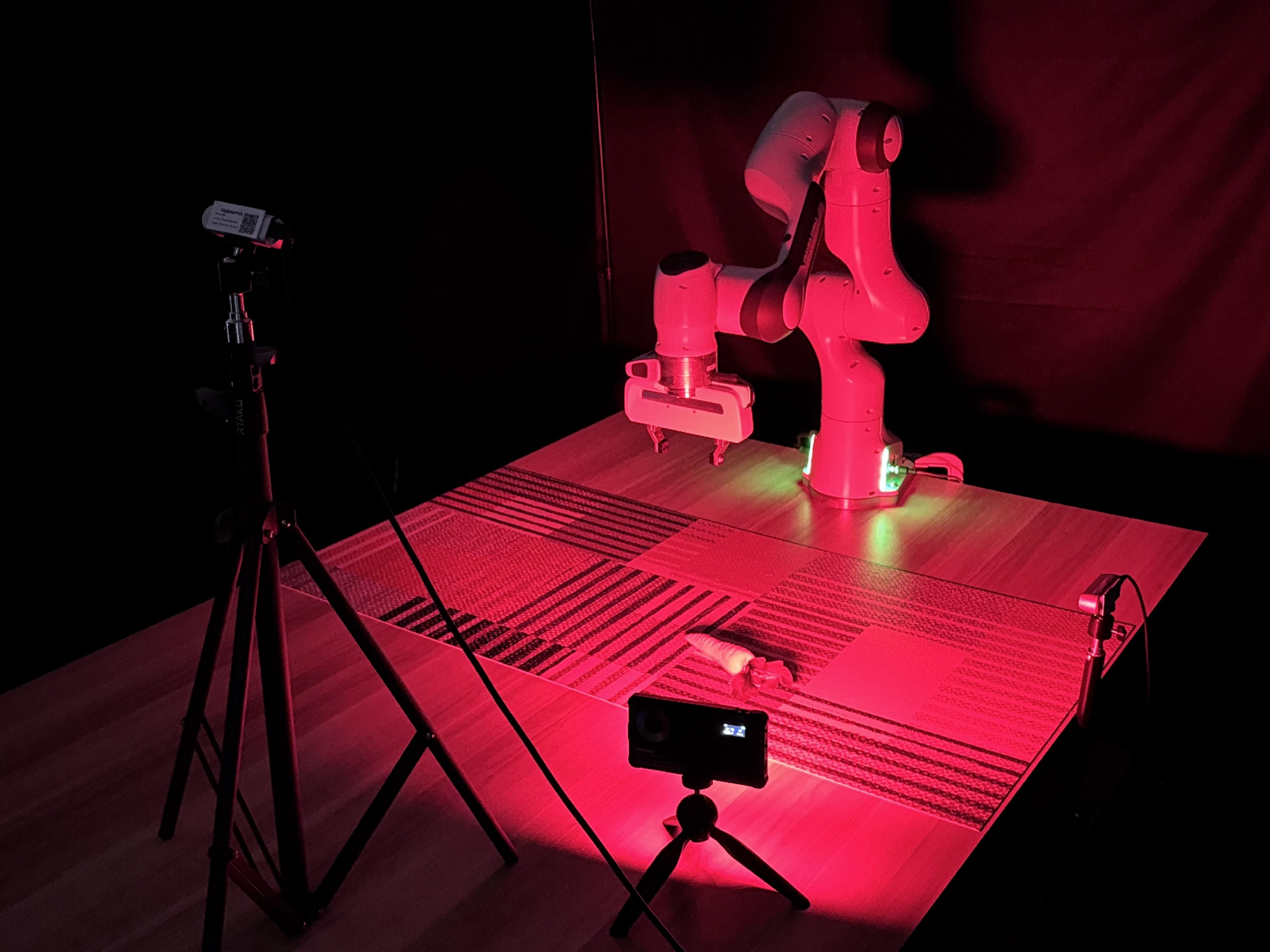}}
    \subfigure[Flashing light (Blue)]{\label{fig:blue_flash}\includegraphics[width=0.3\textwidth]{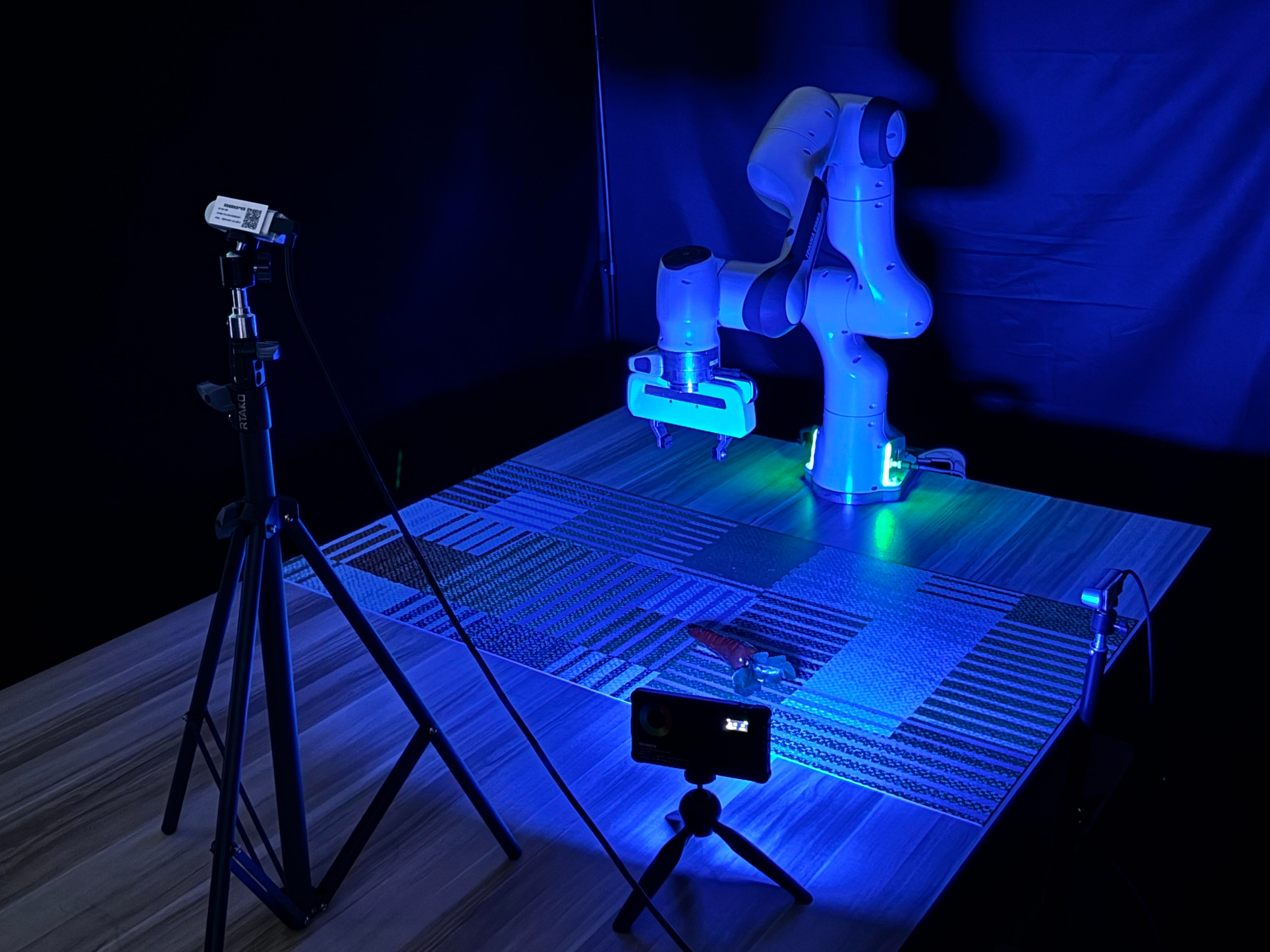}}
    \subfigure[Dark light]{\label{fig:dark}\includegraphics[width=0.3\textwidth]{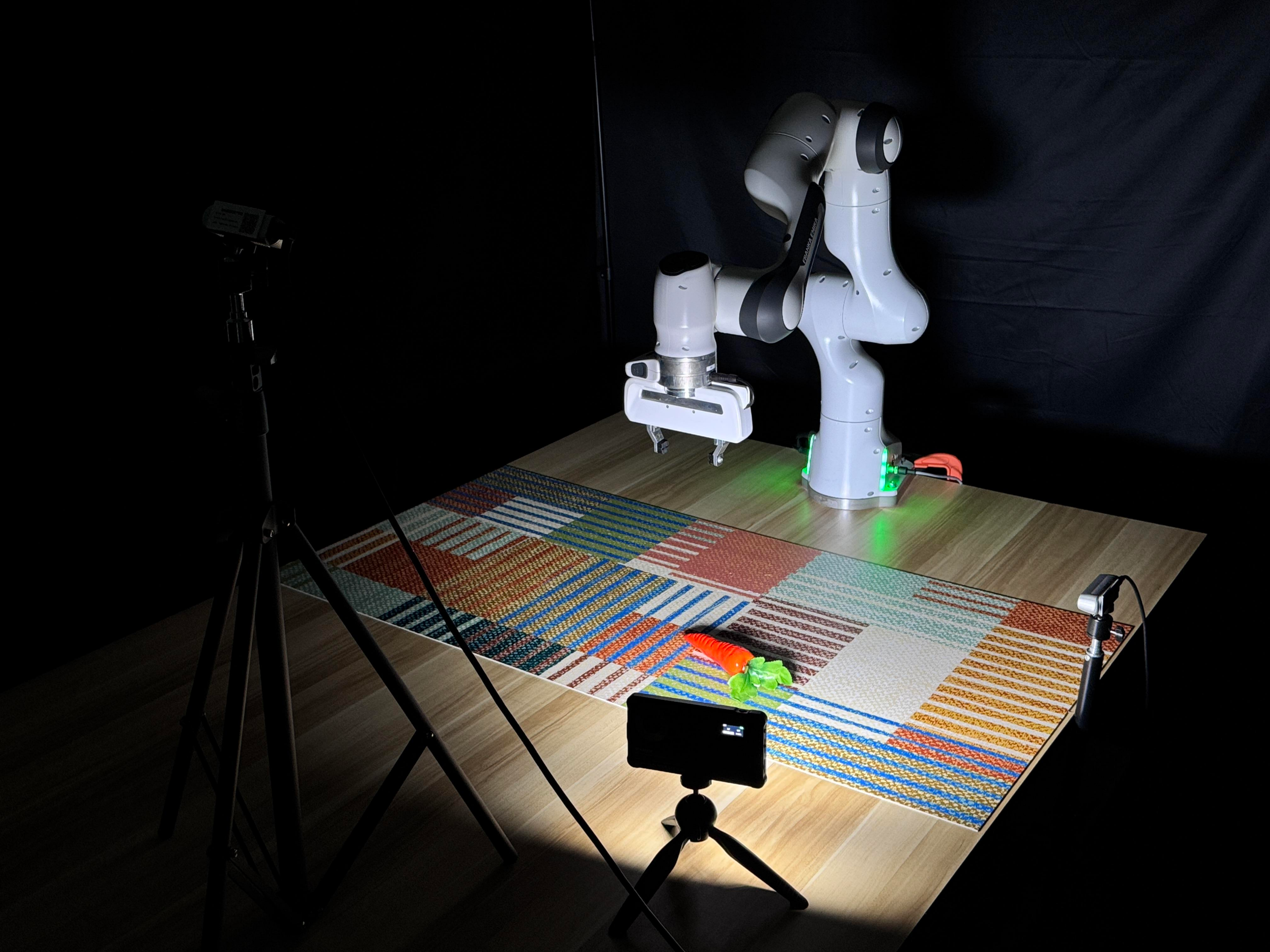}}
    \hfill
    \subfigure[Bright light]{\label{fig:bright}\includegraphics[width=0.3\textwidth]{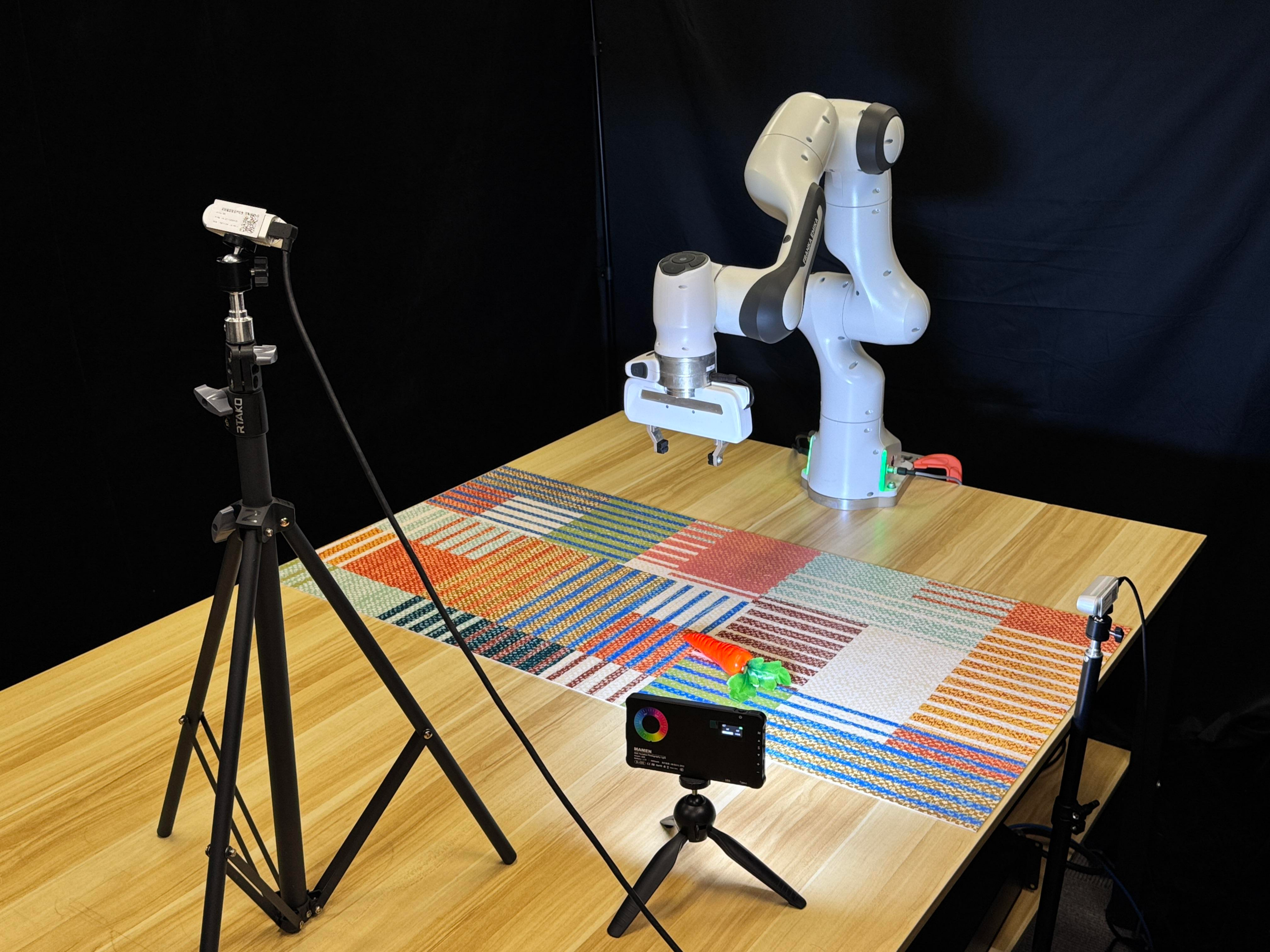}}
    \subfigure[Green light]{\label{fig:green}\includegraphics[width=0.3\textwidth]{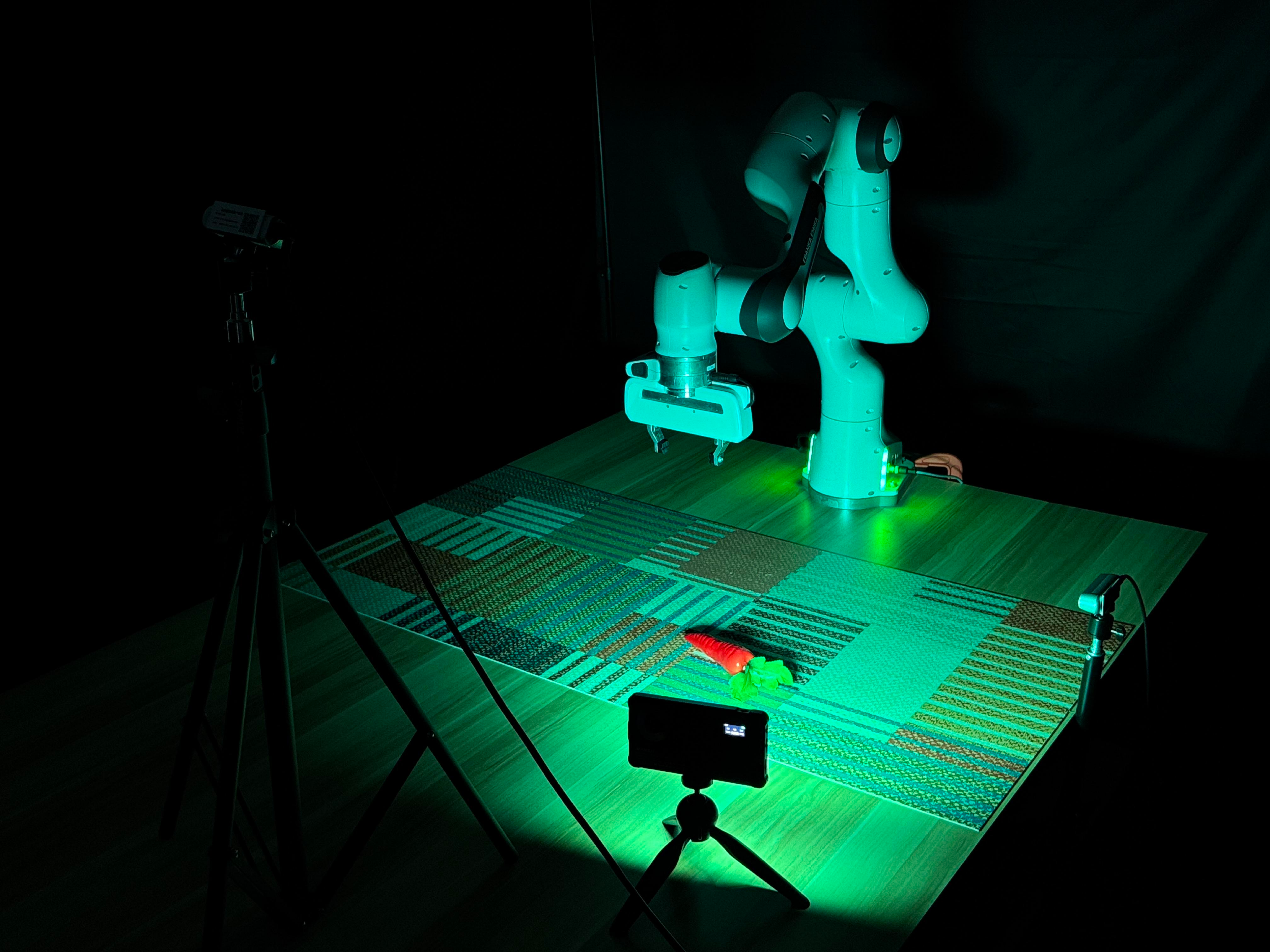}}
    \subfigure[Yellow light]{\label{fig:yellow}\includegraphics[width=0.3\textwidth]{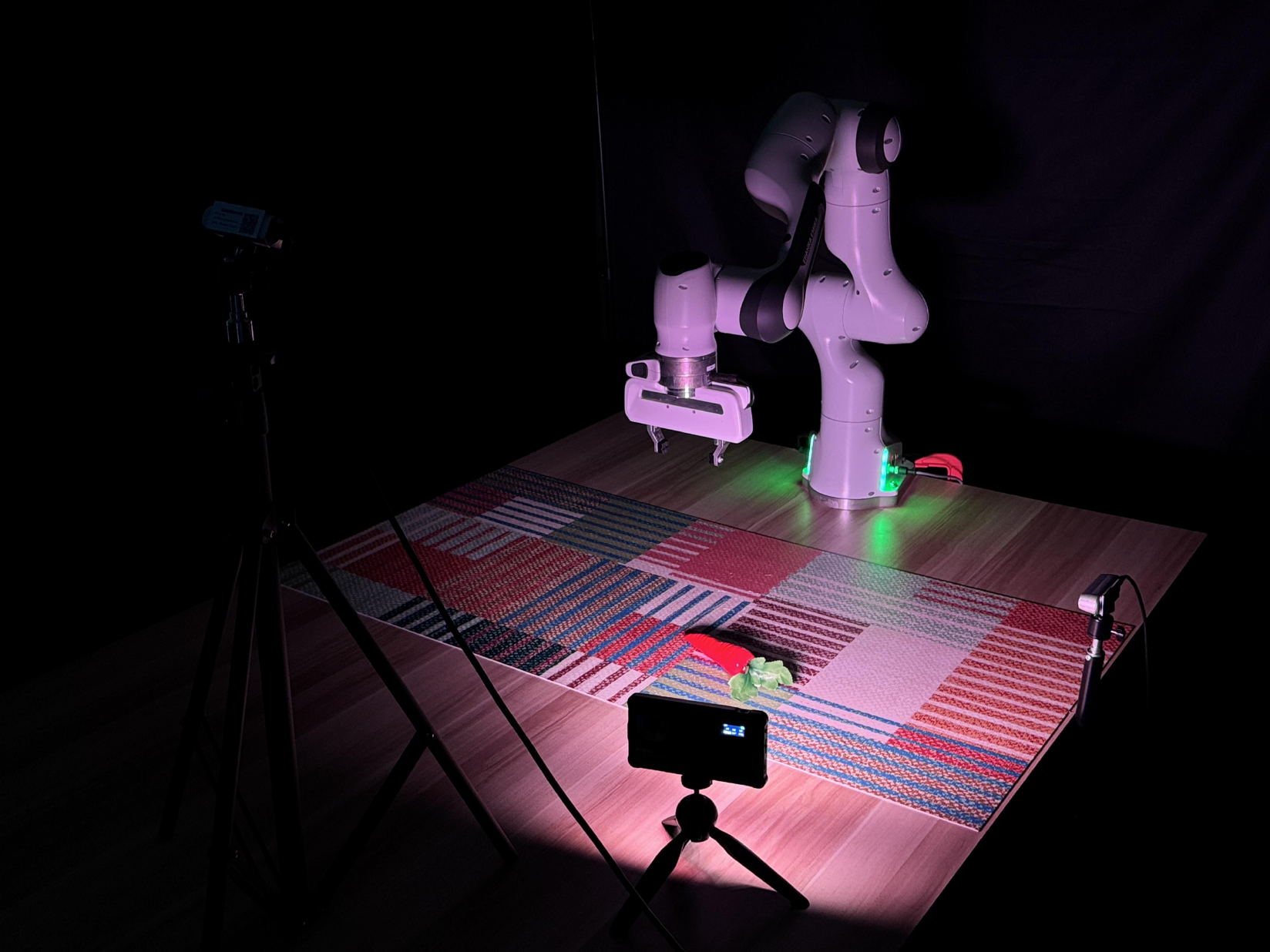}}
    \caption{\textbf{Illustration of real-world experiment on lighting generalization.}}
    \label{fig:real_lighting_demo}
\end{figure*}

We illustrate the experiment settings on lighting condition generalization in Fig.~\ref{fig:real_lighting_demo}.
The flashing light alternates between red and blue light at a frequency of 4Hz.
Every lighting condition takes up 6 trials in a single experiment.

\begin{figure*}[ht]
    \centering
    \subfigure[]{\includegraphics[width=0.3\textwidth]{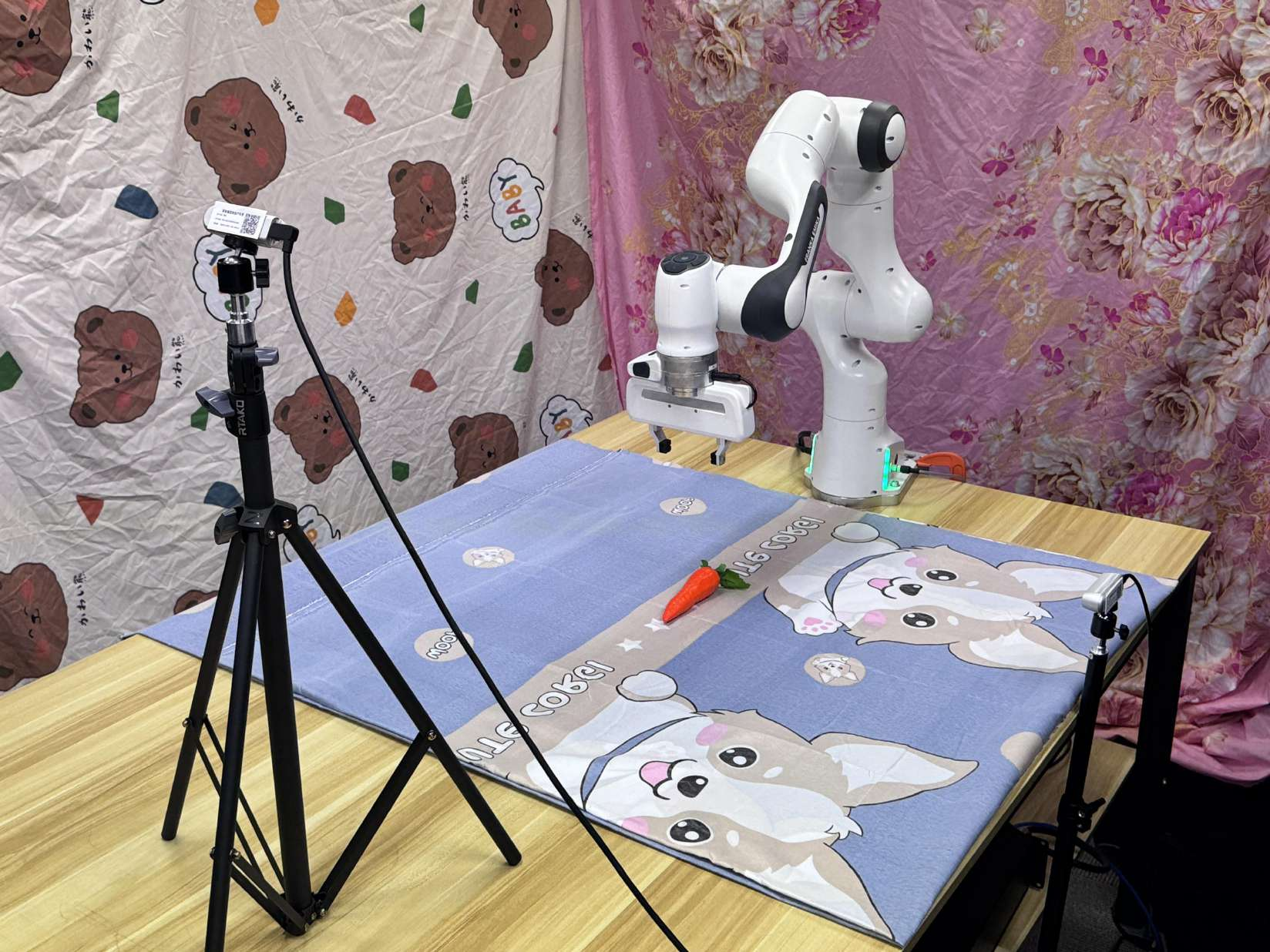}}
    \subfigure[]{\includegraphics[width=0.3\textwidth]{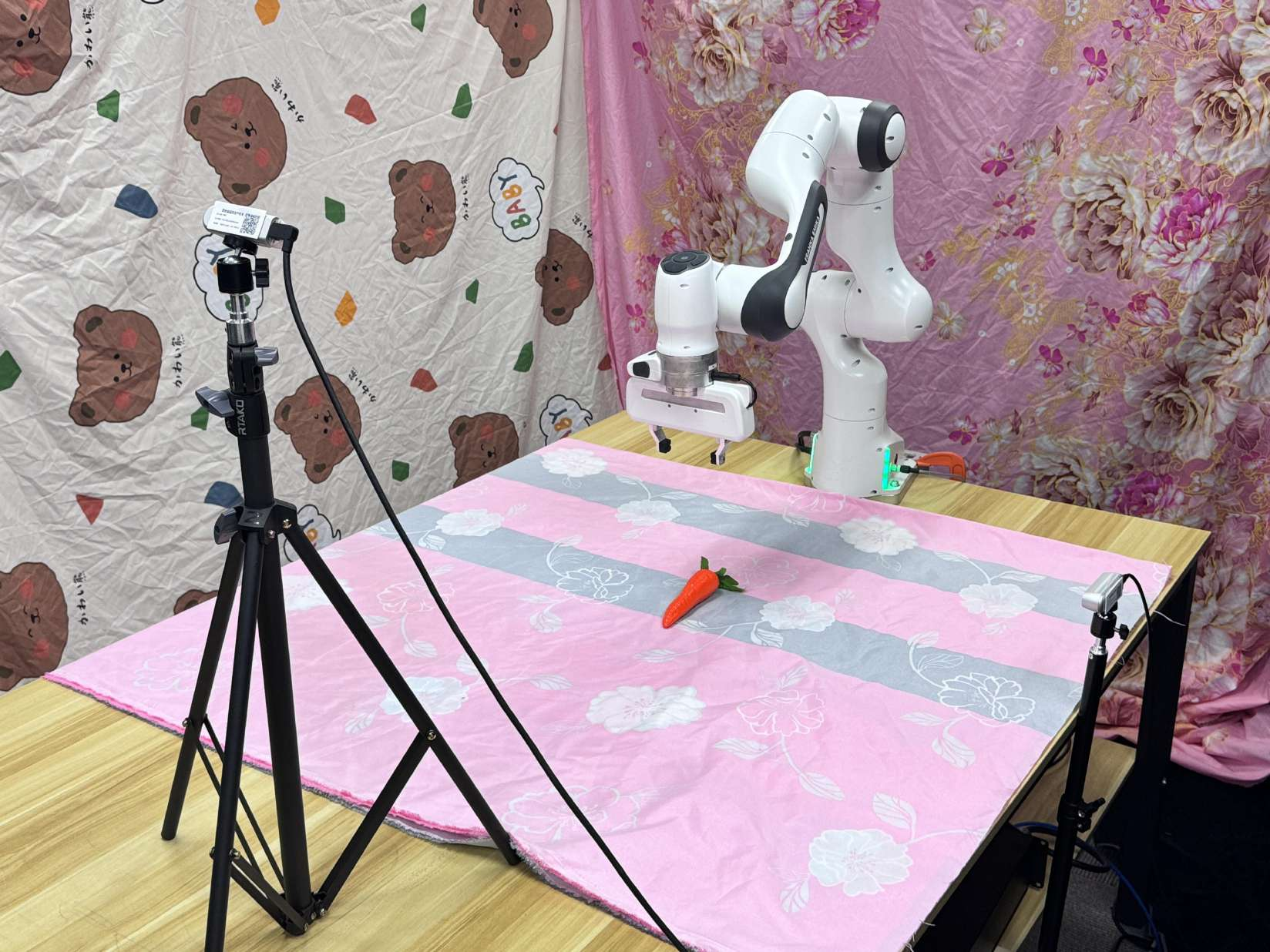}}
    \subfigure[]{\includegraphics[width=0.3\textwidth]{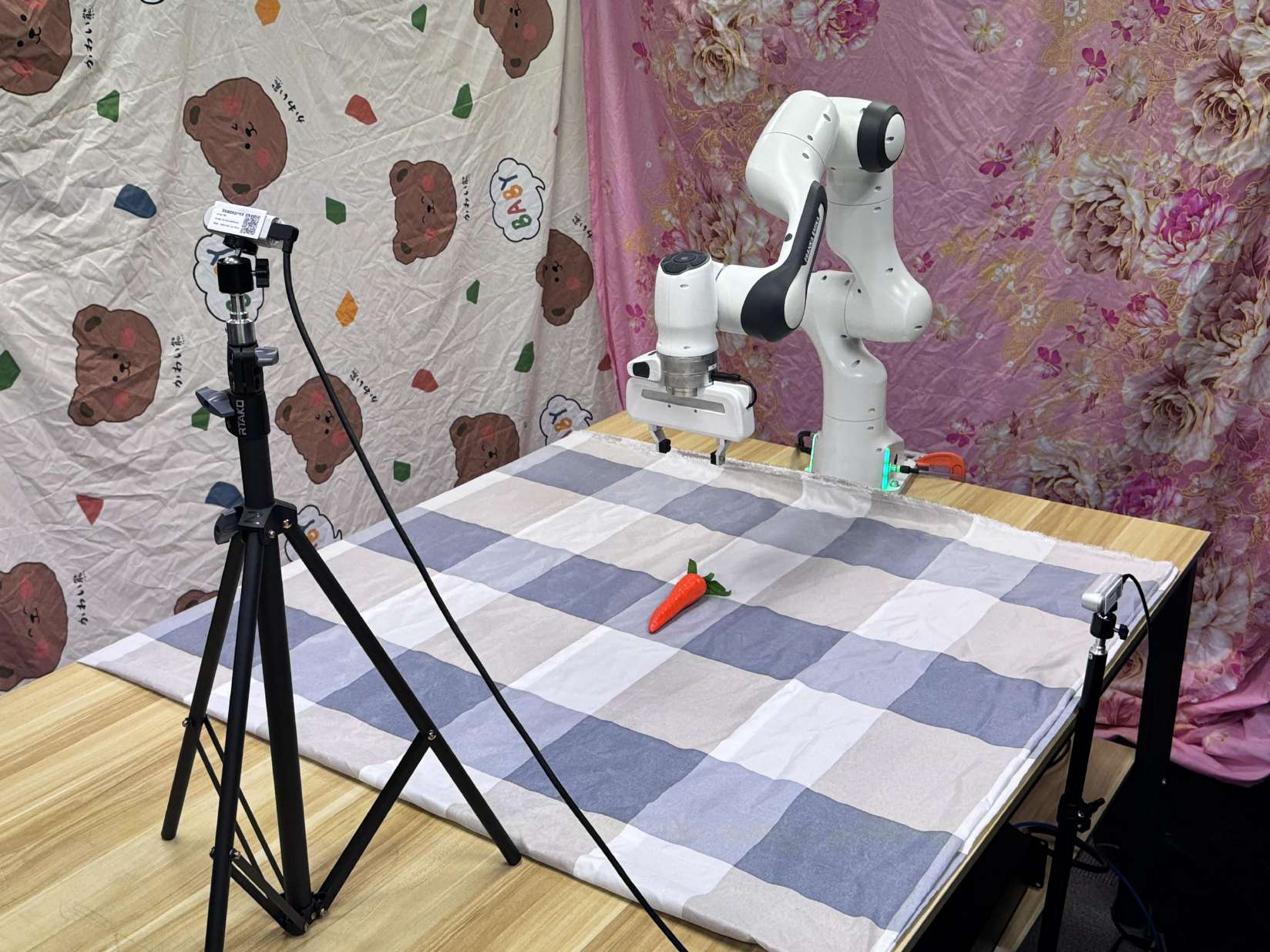}}
    \hfill
    \subfigure[]{\includegraphics[width=0.3\textwidth]{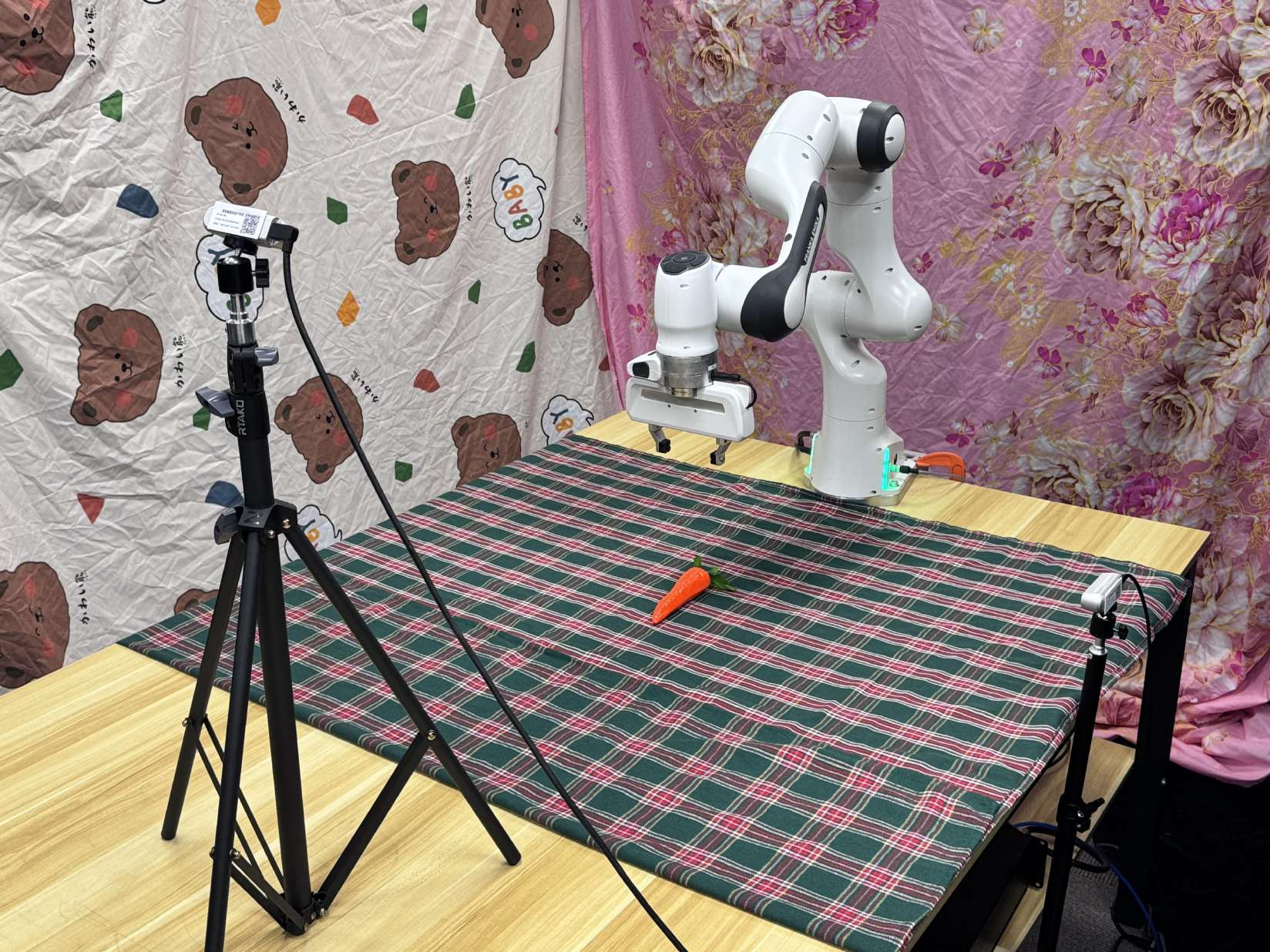}}
    \subfigure[]{\includegraphics[width=0.3\textwidth]{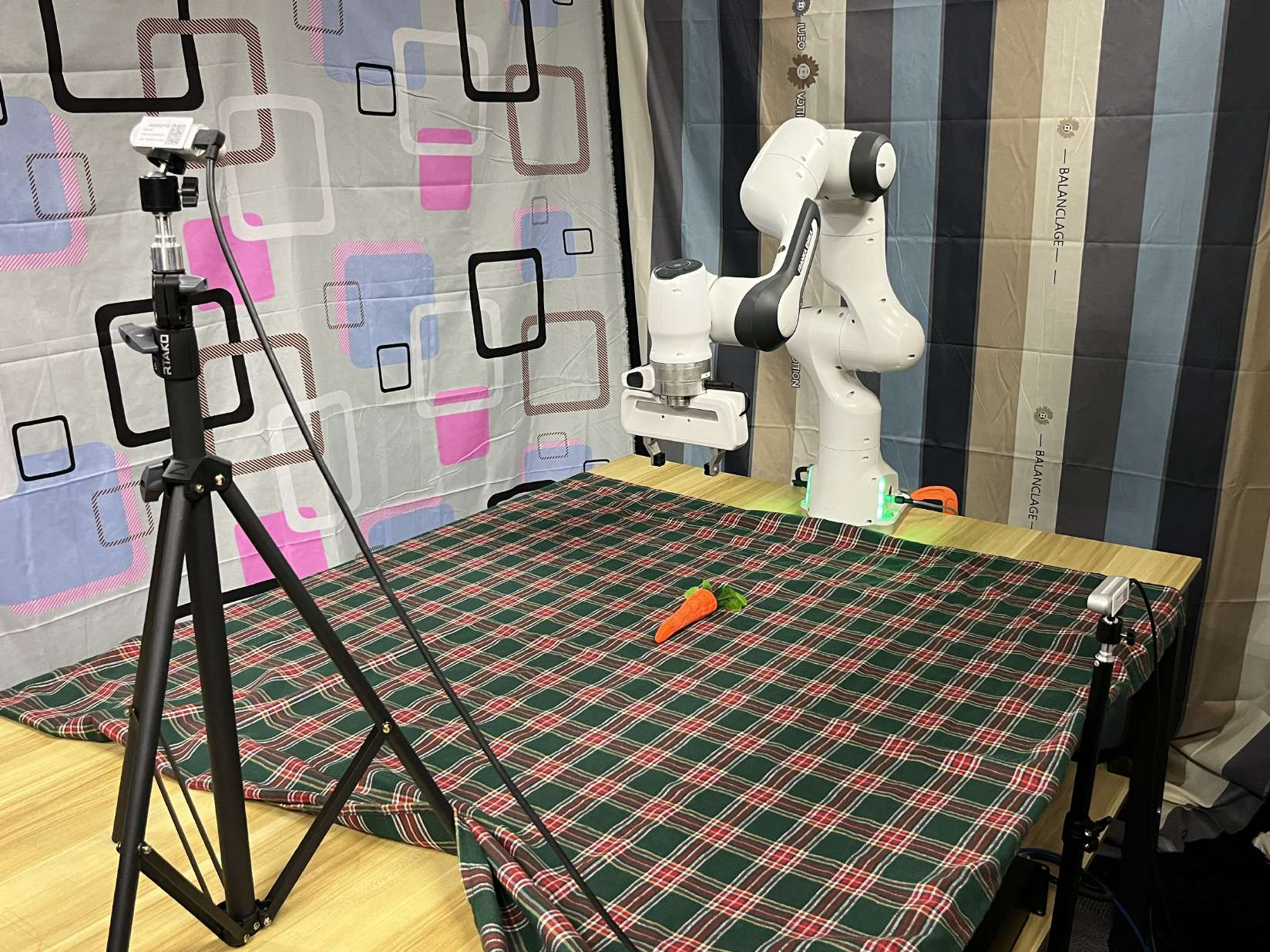}}
    \subfigure[]{\includegraphics[width=0.3\textwidth]{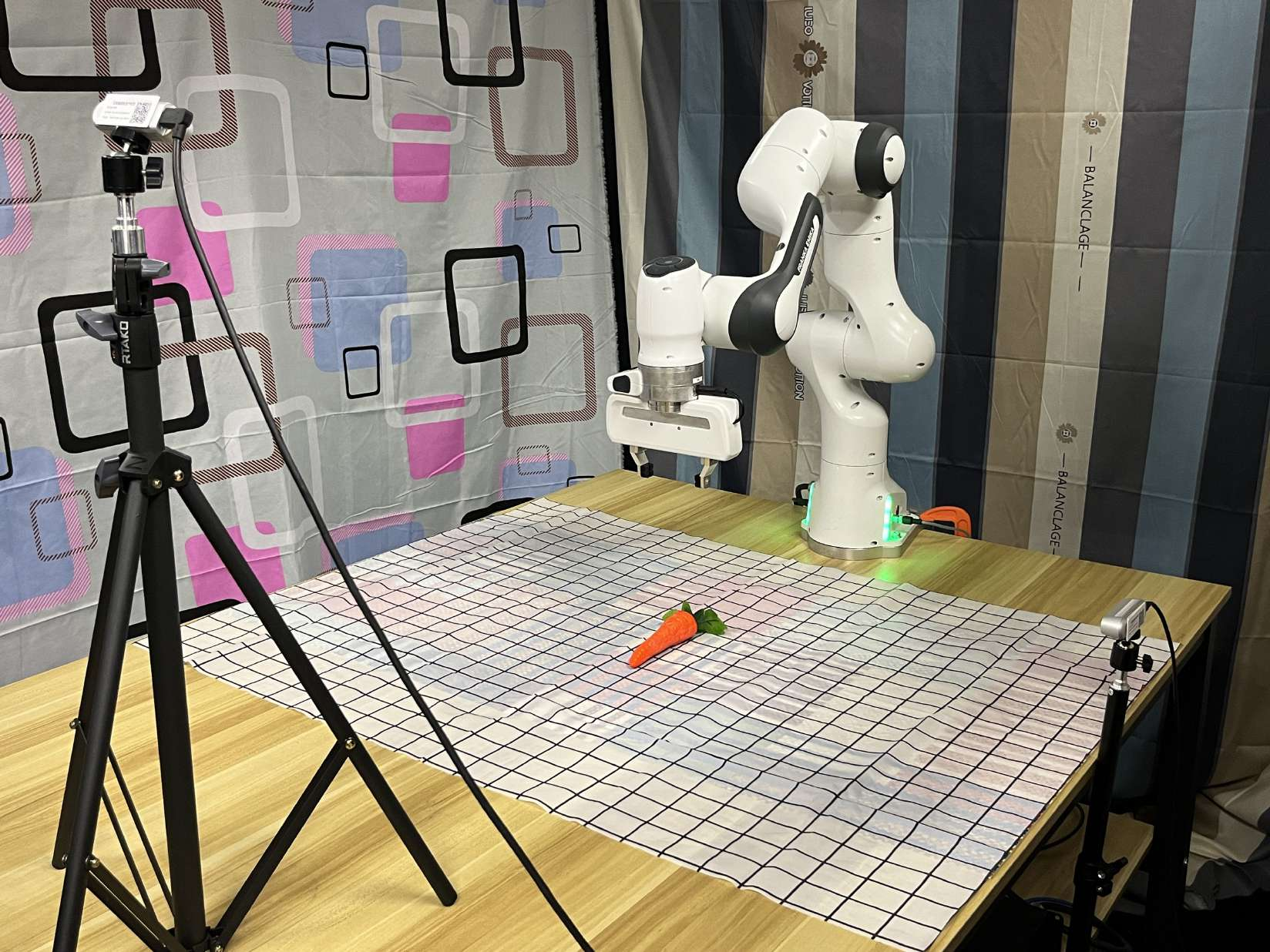}}
    \caption{\textbf{Illustration of real-world experiment on appearance generalization.}}
    \label{fig:real_appearance_demo}
\end{figure*}

Besides, we present the real-world settings on appearance generalization in Fig.~\ref{fig:real_appearance_demo}.
Each scenario accounts for 5 trials in a single experiment.

\end{document}